\documentclass[runningheads]{llncs}
\usepackage{graphicx}
\usepackage{tikz}
\usepackage{comment}
\usepackage{amsmath,amssymb} % define this before the line numbering.
\usepackage{color}

\usepackage[square,numbers]{natbib}
\usepackage[accsupp]{axessibility}  % Improves PDF readability for those with disabilities.

\usepackage{wrapfig}
\newcommand{\etal}{\textit{et al.}}
\usepackage{multirow}
\usepackage{booktabs}
\usepackage{xcolor}
\usepackage{subcaption}
\usepackage{bbm}
\definecolor{ao(english)}{rgb}{0.0, 0.1, 0.0}
\definecolor{slategray}{rgb}{0.44, 0.7, 0.5}
\newcommand{\Drop}[1]{\textcolor{gray}{\textsubscript{\bf $-$#1}}}
\newcommand{\drop}[1]{\textcolor{gray}{\textsubscript{\bf $-$#1}}}
\newcommand{\Rise}[1]{\textcolor{slategray}{\textsubscript{\bf $+$#1}}}

\definecolor{NVblue}{rgb}{0.07,0.12,0.83}
\definecolor{BUred}{rgb}{0.8,0.0,0.0}
\usepackage[pagebackref=true,breaklinks=true,colorlinks=true,citecolor=NVblue,linkcolor=BUred,bookmarks=false]{hyperref}
\usepackage{orcidlink}
\begin{document}

\pagestyle{headings}
\mainmatter
\def\ECCVSubNumber{7719}  % Insert your submission number here

\newcounter{myequation}
\makeatletter
\@addtoreset{equation}{myequation}
\makeatother

\newcounter{myalgorithm}
\makeatletter
\@addtoreset{algorithm}{myalgorithm}
\makeatother

\newcounter{mytable}
\makeatletter
\@addtoreset{table}{mytable}
\makeatother

\newcounter{mysection}
\makeatletter
\@addtoreset{section}{mysection}
\makeatother

\newcounter{myfigure}
\makeatletter
\@addtoreset{figure}{myfigure}
\makeatother

\title{Towards Efficient and Effective Self-Supervised Learning of Visual Representations} % Replace with your title

\begin{comment}
\titlerunning{ECCV-22 submission ID \ECCVSubNumber} 
\authorrunning{ECCV-22 submission ID \ECCVSubNumber} 
\author{Anonymous ECCV submission}
\institute{Paper ID \ECCVSubNumber}
\end{comment}
%******************

% \begin{comment}
\titlerunning{Efficient and Effective Self-Supervised Learning of Visual Representations}
\author{Sravanti Addepalli\thanks{Equal contribution. \newline Correspondence to: Sravanti Addepalli $<$sravantia@iisc.ac.in$>$} \orcidlink{0000-0001-7238-4603},
Kaushal Bhogale$^{\star}$ \orcidlink{0000-0002-0882-9927},
Priyam Dey \orcidlink{0000-0001-5807-1379}, \\ and
R.Venkatesh Babu \orcidlink{0000-0002-1926-1804}}
\authorrunning{S. Addepalli et al.}
\institute{Video Analytics Lab, Department of Computational and Data Sciences, \\Indian Institute of Science, Bangalore}
% \end{comment}
%******************
\maketitle

\begin{abstract}
Self-supervision has emerged as a propitious method for visual representation learning after the recent paradigm shift from handcrafted pretext tasks to instance-similarity based approaches. Most state-of-the-art methods enforce similarity between various augmentations of a given image, while some methods additionally use contrastive approaches to explicitly ensure diverse representations. While these approaches have indeed shown promising direction, they require a significantly larger number of training iterations when compared to the supervised counterparts. In this work, we explore reasons for the slow convergence of these methods, and further propose to strengthen them using well-posed auxiliary tasks that converge significantly faster, and are also useful for representation learning. The proposed method utilizes the task of rotation prediction to improve the efficiency of existing state-of-the-art methods. We demonstrate significant gains in performance using the proposed method on multiple datasets, specifically for lower training epochs.

\end{abstract}

\section{Introduction}

\label{sec:intro}
The unprecedented progress achieved using Deep Neural Networks over the past decade was fuelled by the availability of large-scale labelled datasets such as ImageNet \citep{imagenet_cvpr09}, coupled with a massive increase in computational capabilities.
While their initial success was contingent on the availability of annotations in a supervised learning framework \citep{alexnet, inception, resnet, lecun2015deep}, recent years have witnessed a surge in self-supervised learning methods, which could achieve comparable performance, albeit using a higher computational budget and larger model capacities \cite{simclr,simsiam,byol,swav}. 
Early self-supervised approaches \cite{colorization,jigsaw,rotnet} aimed at learning representations while solving specialized tasks that require a semantic understanding of the content to accomplish. While generative networks such as task-specific encoder-decoder architectures \citep{kingma2013auto,zhang2017split,inpainting} and Generative Adversarial Networks (GANs) \citep{goodfellow2014generative,donahue2016adversarial} could learn useful representations, they were superseded by the use of discriminative tasks such as solving Jigsaw puzzles \citep{jigsaw} and rotation prediction \citep{rotnet}, as the latter could be achieved using lower model capacities and lesser compute. The surprisingly simple task of rotating every image by a random angle from the set $\{0^{\circ},90^{\circ},180^{\circ},270^{\circ}\}$, and training the network to predict this angle was seen to outperform other handcrafted task based methods with a similar convergence rate as supervised training \cite{rotnet}. Compared to these pretext task based methods, recent approaches have achieved a significant boost in performance by learning similar representations across various augmentations of a given image \citep{moco, byol, simclr,swav,simsiam}. While these methods show improvements at a low training budget as well, they achieve a further boost when trained for a larger number of epochs \cite{simsiam}, indicating that improving the convergence of such methods can lead to valuable gains at a low computational cost. 

\begin{figure}[t]
\centering
\includegraphics[width=\linewidth]{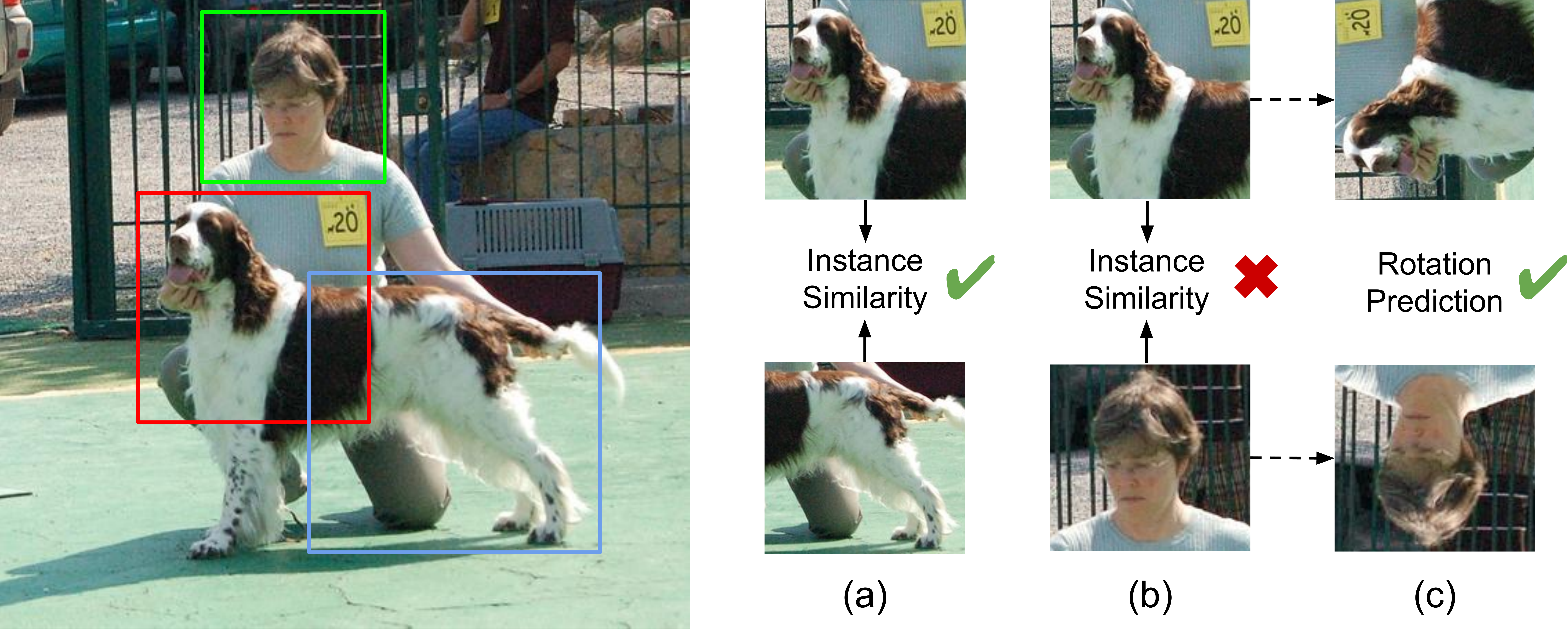}
\caption{\small{We demonstrate noise in the training objective of instance-similarity based learning tasks. Consider the three random crops shown in the input image. The two crops in (a) are desirable, while the crops shown in (b) give an incorrect signal to the network. Since the task of rotation prediction shown in (c) aims to predict the rotation angle of each cropped image independently, there is no noise in the training objective.}}
\label{fig:noise_learning}
\end{figure}

In this work, we empirically show that a key reason for the slow convergence of instance-similarity based approaches is the presence of noise in the training objective, owing to the nature of the learning task, as shown in Fig.\ref{fig:noise_learning}. We further propose to strengthen the recent state-of-the-art instance-similarity based self-supervised learning algorithms such as BYOL \citep{byol} and SwAV \cite{swav} using a noise-free auxiliary training objective such as rotation prediction in a multi-task framework. As shown in Fig.\ref{fig:acc_cifar_all}, this leads to a similar convergence rate as RotNet \cite{rotnet}, while also resulting in better representations from the instance-similarity based objective. We further study the invariance of the network to geometric transformations, and show that in natural images, rotation invariance hurts performance and learning covariant representations across multiple rotated views leads to improved results. We demonstrate significant gains in performance across multiple datasets - CIFAR-$10$, CIFAR-$100$ \citep{krizhevsky2009learning} and ImageNet-100 \citep{cmc,imagenet_cvpr09}, and the scalability of the proposed approach to ImageNet-1k \cite{imagenet_cvpr09} as well.

\noindent Our code is available here: \url{https://github.com/val-iisc/EffSSL}. 

\section{Related Works}

\subsection{Handcrafted Pretext task based methods}

Discriminative pretext tasks use pseudo-labels that are generated automatically without the need for human annotations. This includes tasks based on spatial context of images such as context prediction \citep{context}, image jigsaw puzzle \citep{jigsaw} and counting visual primitives \citep{counting}. 

\textbf{RotNet:} Rotation prediction, proposed by Gidaris et al. \cite{rotnet}, has been one of the most successful pretext tasks for the learning of useful semantic representations.  In this approach, every image is transformed using all four rotation transformations, and the network is trained to predict the corresponding rotation angle used for transforming the image.
Due to its simplicity and effectiveness, the rotation task has been used to improve the training of GANs \citep{goodfellow2014generative,chen2019self} as well.

\textbf{Multi-task Learning:} Doersch and Zisserman \cite{doersch2017multi} investigated methods for combining several pretext tasks in a multi-task learning framework to learn better representations. Contrary to a general multi-task learning setting, in this work we aim to improve instance similarity based tasks such as BYOL \citep{byol} and SwAV \citep{swav} using handcrafted pretext tasks. We empirically show that the training objective of instance-similarity based tasks is noisy, and combining it with the well-defined objective of rotation prediction leads to improved performance.

\subsection{Instance Discriminative approaches}

Recent approaches aim to learn similar representations for different augmentations of the same image, while generating diverse representations across different images. Several works achieve this using contrastive learning approaches \citep{cpc, datacpc, simclr, moco, pirl}, where multiple augmentations of a given image are considered as positives, and augmentations of other images are considered as negatives. PIRL \cite{pirl} and MoCo \cite{moco} maintain a queue to sample more negatives.

\textbf{SimCLR:} The work by Chen et al. \cite{simclr} presents a Simple Framework for Contrastive Learning of Visual Representations (SimCLR), that utilizes existing architectures such as ResNet \citep{resnet}, and avoids the need for memory banks. The authors proposed the use of multiple data augmentations and a learnable nonlinear transformation between representations to improve the effectiveness of contrastive learning. 
Two independent augmentations for every image are considered as positives in the contrastive learning task, while the augmentations of all other images are considered as negatives. 
The network is trained by minimizing the normalized temperature-scaled cross entropy loss (NT-Xent) loss.

\textbf{BYOL, SimSiam:} While prior approaches relied on the use of negatives for training, Grill et al. \cite{byol} proposed Bootstrap Your Own Latent (BYOL), which could achieve state-of-the-art performance without the use of negatives. The two augmentations of a given image are passed through two different networks - the base encoder and the momentum encoder. 
The base encoder is trained such that the representation at its output can be used to predict the representation at the output of the momentum encoder, using a predictor network. Chen and He \cite{simsiam} show that it is indeed possible to avoid a collapsed representation even without the momentum encoder using Simple Siamese (SimSiam) networks, and that the stop-gradient operation is crucial for achieving this. 

\textbf{Clustering based methods, SwAV:} Clustering-based self-supervised approaches use pseudo-labels from the clustering algorithm to learn representations. DeepCluster \cite{caron2018deep} alternates between using k-means clustering for producing pseudo-labels, and training the network to predict the same. Asano \etal \cite{asano2019self} show that degenerate solutions exist in the DeepCluster \cite{caron2018deep} algorithm. To address this, they cast the pseudo-label assignment problem as an instance of the optimal transport problem and solve it efficiently using a fast variant of the Sinkhorn-Knopp algorithm \cite{cuturi2013sinkhorn}. SwAV \citep{swav} also uses the Sinkhorn-Knoop algorithm for clustering the data while simultaneously enforcing consistency between cluster assignments by Swapping Assignments between Views (SwAV), and using them as targets for training.

\subsection{Relation with concurrent works} There has been some recent interest towards improving instance-similarity based approaches by combining them with pretext tasks \citep{kinakh2021scatsimclr,dangovski2021equivariant} . In particular, Kinakh et al. \cite{kinakh2021scatsimclr} show that the use of pretext auxiliary tasks in addition to the contrastive loss can boost the accuracy of models like ScatNet and ResNet-18 on small-scale datasets like STL-10 and CIFAR-100-20. Dangovski et al. \cite{dangovski2021equivariant} claim that learning equivariant representations is better than learning invariant representations, and hence the auxiliary rotation prediction task helps. 
Our work complements these efforts, and highlights a key issue in the instance-discriminative learning objective: the \textit{impact of noise} in their slow convergence, and shows that combining them with a \textit{noise-free} auxiliary pretext task can significantly improve their efficiency and effectiveness.

\section{Motivation}
\label{issues}

The evolution of self-supervised learning algorithms from handcrafted pretext task-based methods \citep{colorization, rotnet, jigsaw} to instance discriminative approaches \citep{moco, byol, simclr,swav,simsiam} has indeed led to a significant boost in the performance of downstream tasks. However, as shown in Fig.\ref{fig:acc_cifar_all}, the latter require a larger number of training epochs for convergence. In this section, we show using controlled experiments that the slow convergence of instance-discriminative algorithms can be attributed to a noisy training objective, and eliminating this noise can lead to improved results. 

\begin{table}
\begin{minipage}{0.48\linewidth}
\caption{\textbf{Eliminating False Negatives} in contrastive learning across varying levels of supervision ($\%$ Labels). Elimination of noise in the training objective leads to higher linear evaluation accuracy ($\%$) within a fixed training budget.}

\small
\centering
\label{tab:FN_simclr}
\resizebox{1.0\linewidth}{!}{
\begin{tabular}{cllc}
\toprule
$\%$ ~Labels~ & SimCLR~& Ours~  & ~Gain ($\%$)~ \\
\midrule
0                                    & 88.77                    & 90.91                         & 2.14    \\
30                                   & 92.26 \Rise{3.49}                    & 93.94                          & 1.68    \\
50                                   & 92.93 \Rise{0.67}                   & 94.02                          & 1.09    \\
100                                  & 93.27 \Rise{0.34}                   & 94.15                         & 0.88   \\
\bottomrule
\end{tabular}}
\end{minipage}
\hfill
\begin{minipage}{0.48\linewidth}
\caption{\textbf{Eliminating False Positives} in BYOL \cite{byol} across varying levels of supervision ($\%$ Good Crops). Elimination of noise in the training objective leads to higher linear evaluation accuracy ($\%$) within a fixed training budget.}

\small
\centering
\label{tab:FP_byol}
\resizebox{1.0\linewidth}{!}{
\begin{tabular}{cllc}
\toprule
\% Good Crops & BYOL  & Ours& Gain ($\%$) \\
\midrule
0   & 63.64 & 68.62 & 4.98 \\
25  & 64.50\Rise{0.86} & 68.30 & 3.80 \\
50  & 66.30\Rise{1.80} & 68.90 & 2.60 \\
100 & 66.72\Rise{0.42} & 70.26 & 3.54 \\
\bottomrule
\end{tabular}}

\end{minipage}

\end{table}

\subsection{Impact of False Negatives in SimCLR} 
\label{subsec:FN_simclr}

The contrastive learning objective in SimCLR \cite{simclr} considers two augmentations of a given image as positives and the augmentations of all other images in the batch as negatives. These negatives could belong to the same class as the anchor image, and possibly be as similar to the anchor image as the corresponding positive, leading to a noisy training objective. While the probability of same class negatives is higher when batch size is higher than the number of classes, this issue can occur even otherwise, when there exist negative images that are more similar to the anchor when compared to the positive. Khosla et al. \cite{khosla2020supervised} use supervision from labels in a Supervised Contrastive (SupCon) framework to convert the same-class false negatives to additional positives, and show an improvement over supervised learning methods. 

In order to specifically study the impact of eliminating false-negatives, we first perform experiments by using labels to avoid using the same class samples as negatives. We do not add these eliminated negatives as positives, in order to avoid excessive supervision. In Table-\ref{tab:FN_simclr} we present results of an experiment on the CIFAR-$10$ dataset, where the same-class negatives in SimCLR are eliminated using a varying fraction of labels. The fraction of labels serves as an upper bound to the amount of noise reduction in the training objective, considering that other sources of noise such as false-positives are still not eliminated. Using $30\%$ labels, we achieve a $3.49\%$ increase in accuracy when compared to the SimCLR baseline ($0\%$ labels case). It is also interesting to note that the boost in accuracy is highest for $30\%$ supervision and reduces as the fraction of labels increase. This indicates that the network can possibly overcome the impact of noise more effectively when the amount of noise is lower. Overall, we obtain $4.5\%$ boost in the case where all the labels are used. By jointly training SimCLR with the task of rotation prediction (Ours), we achieve highest gains in the case of $0\%$ labels or the no supervision case, and significantly lower gains as the amount of labels increase. We discuss this in greater detail in Section-\ref{sec:rob_noise}.

\subsection{Impact of False Positives in BYOL} 
\label{subsec:FP_byol}
Since BYOL does not use a contrastive learning objective, it is not directly impacted by noise due to false negatives. However, as shown in Fig.\ref{fig:noise_learning}(b), the augmentations considered may not be similar to each other, leading to false positives. Selvaraju et al. \cite{selvaraju2020casting} show that unsupervised saliency maps can be used for the selection of better crops, and also as a supervisory signal in the training objective. This leads to improved performance on scene datasets which contain multiple objects. 
Inspired by this, we use Grad-CAM \citep{gradcam}  based saliency maps from a supervised ImageNet pre-trained network to select crops such that the ratio of mean saliency score of the cropped image and that of the full image is higher than a certain threshold (Details in Sec.\ref{FP_supp}). It is to be noted that the only difference with respect to BYOL is in the use of supervised saliency maps for the selection of crops. Alternatively, unsupervised saliency maps could also be used for the same. We demonstrate results of this experiment on the ImageNet-100 dataset in Table-\ref{tab:FP_byol}. We observe that by using saliency-maps for crop selection, the accuracy improves by $3.08\%$ for a fixed training budget. While this experiment shows the impact of reducing the false positives in the BYOL objective, it does not completely eliminate noise in the training objective, since the saliency maps themselves are obtained from a Deep Neural Network, and hence may not be very accurate.

\section{Proposed Method}

In this section, we examine the advantages of instance-discriminative approaches and handcrafted pretext-task based methods, and further discuss our proposed approach which integrates both methods to overcome their limitations.

The key ingredients for the success of a self-supervised learning algorithm are (i) Well-posedness of the learning task; (ii) Extent of correlation between representations that help accomplish the pretext task, and ideal representations, whose quality is evaluated using downstream tasks.

The success of instance-similarity based approaches in achieving state-of-the-art performance on downstream tasks indeed shows that the representations learnt using such tasks are well correlated with ideal representations. However, these methods require to be trained on a significantly larger number of training epochs when compared to the supervised counterparts. As seen in the previous section, a possible reason for the slow convergence of these methods is the noise in training objective due to the presence of false positives and false negatives. While it is possible to overcome this noise using additional supervision from (unsupervised) pre-trained models, such as the use of saliency maps for crop selection, these methods are not very successful as this supervisory signal is also not perfect in practice. Moreover, this method assumes the availability of a network which is pre-trained on a relevant dataset, which may not always hold true, and hence adds to the training cost. We observe that the boost in performance is not good enough to justify the additional computational overhead. If the same computational budget is invested in the main self-supervised task, it leads to better performance (Details in Sec.\ref{FP_supp}).

 \begin{figure}
     \centering
    \includegraphics[width=0.6\linewidth]{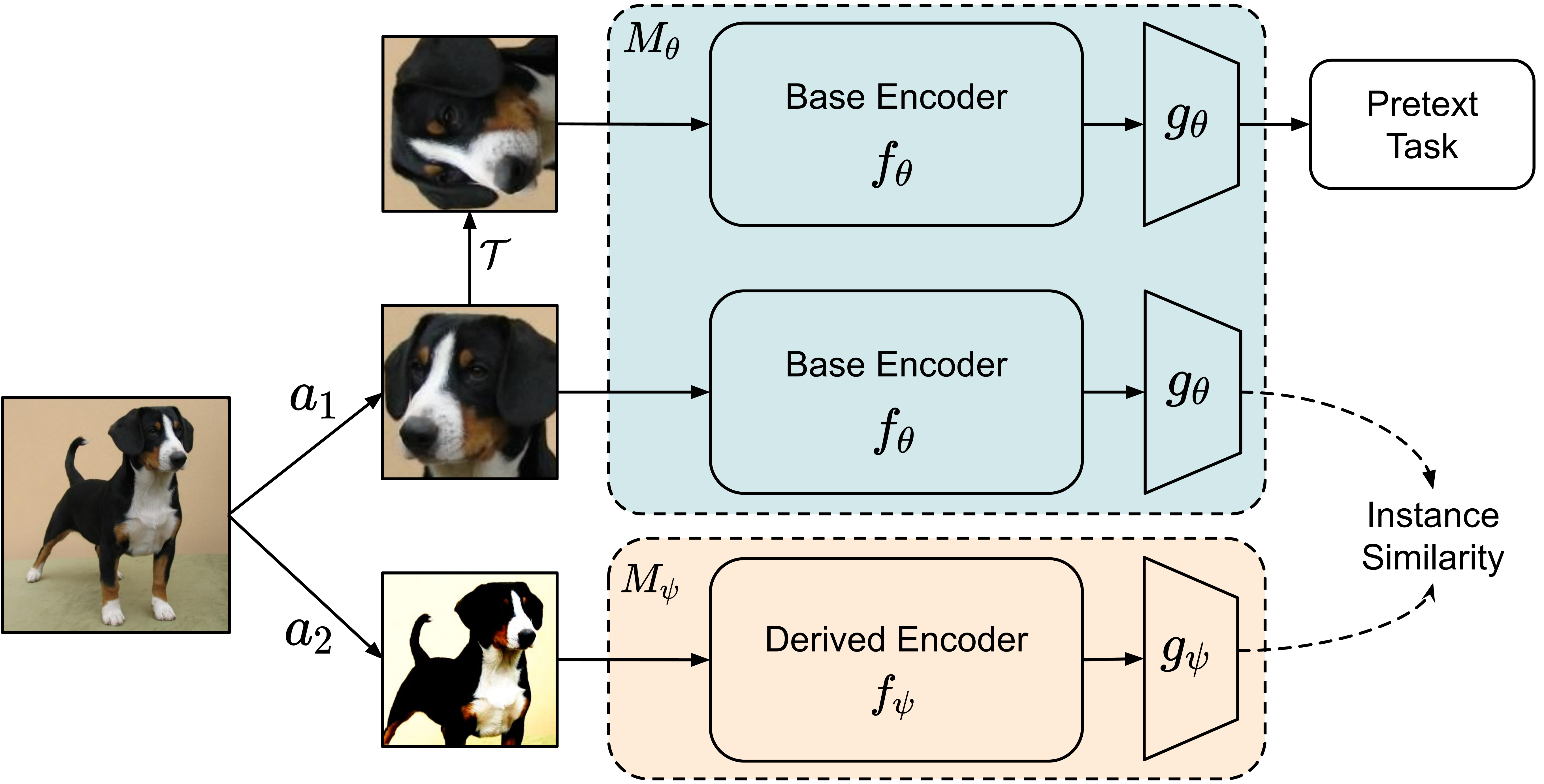}
    \caption{\small{Schematic diagram illustrating the proposed approach. A pretext task such as rotation prediction is combined with base methods like BYOL and SimCLR. For methods like BYOL and MoCo, the derived network $M_\psi$ is a momentum-averaged version of $M_\theta$, and for methods like SimCLR, $M_\theta$ and $M_\psi$ share the same parameters.}}
     \label{fig:block_diagram}
 \end{figure}

On the other hand, task-based objectives such as rotation prediction score higher on the well-posedness of the learning task. In this task, a known random rotation transformation is applied to an image, and the task of the network is to predict the angle of rotation. Since the rotation angle is known a priori, there is very little scope for noise in labels or in the learning objective, leading to faster training convergence. 

In this work, we propose to enhance the convergence of instance-similarity based approaches using pretext-task based objectives such as rotation prediction. The proposed approach can be used to enhance many existing instance-discrimination based algorithms (referred to as base algorithm) as shown in Section-\ref{sec:exp}. A schematic diagram of our proposed approach is presented in Fig.\ref{fig:block_diagram}. 

We term the main feature extractor to be learned as the base encoder, and denote it as $f_\theta$. Some of the self-supervised learning algorithms use an additional encoder, which is derived from the weights of the base encoder. We call this as a derived encoder and represent it using ${f_\psi}$. It is to be noted that the derived encoder may be also be identical to the base encoder, which represents an identity mapping between $\theta$ and $\psi$. As proposed by Chen et al. \cite{simclr}, many of the approaches use a learnable nonlinear transformation between the representations and the final instance-discriminative loss. We denote this projection network and its derived network using $g_\theta$ and $g_\phi$ respectively. We note that the base algorithm may have additional layers between the projection network and the final loss, such as the predictor in BYOL \cite{byol} and SimSiam \cite{simsiam}, which are not explicitly shown in Fig.\ref{fig:block_diagram}. 
 
An input image $x$ is first subject to two augmentations $a_1$ and $a_2$ to generate $x^{a_1}$ and $x^{a_2}$. We use the augmentation pipeline from the respective base algorithm such as BYOL or SimCLR. These augmented images are passed through the base encoder $f_\theta$ and the derived encoder $f_\phi$ respectively, and the outputs of the projection networks $g_\theta$ and $g_\phi$ are used to compute the training objective of the respective base algorithm. The augmentation $x^{a_1}$ is further transformed using a rotation transformation $t$ which is randomly sampled from the set $\mathcal{T} = \{0^{\circ},90^{\circ},180^{\circ},270^{\circ}\}$. The rotated image $x^{a_1,t}$ is passed through the base encoder $f_\theta$ and projection network $g_\theta$ which are shared with the instance-based task. We represent the overall network formed by the composition of $f_\theta$ and $g_\theta$ by $M_\theta$, and similarly the composition of $f_\psi$ and $g_\psi$ by $M_\psi$. The representation $M_\theta(x^{a_1,t})$ is input to a task-specific network $h_\theta$ whose output is a $4$-dimensional softmax vector over the outputs in the set $\mathcal{T}$. The overall training objective is as follows:
\begin{equation}
\label{eq:ours}
    \mathcal{L} = \mathcal{L}_\textrm{base} + \lambda \cdot \frac{1}{2B}\sum_{i=0}^{B-1} \sum_{m=1}^{2} \ell_{CE}(h_\theta(M_\theta(x^{a_m,t_k}_i),t_k))
\end{equation}

Here $t_k$ is sampled uniformly at random for each image from the set $\mathcal{T}$. $\mathcal{L}_\textrm{base}$ represents the symmetric loss of the base instance-similarity based algorithm used. We describe the base loss for BYOL \cite{byol} and SimCLR \cite{simclr} in Sec.\ref{background}. $\lambda$ is the weighting factor between rotation task and the instance-similarity objective. While the RotNet algorithm \cite{rotnet} uses all four rotations for every image, we consider only two in the overall symmetric loss. Therefore, when compared to the base algorithm, the computational overhead of the proposed method is limited to one additional forward propagation for every augmentation, which is very low when compared to the other components of training such as data loading and backpropagation. There is no additional overhead in backpropagation since the combined loss (Eq.\ref{eq:ours}) is used for training.

\section{Experiments and Analysis}
\label{sec:exp}

In this section, we first describe our experimental settings (Sec.\ref{exp_setup}), following which we present an empirical analysis to highlight the importance of the auxiliary task towards improving the efficiency and effectiveness of the base learning algorithm (Sec.\ref{sec:rob_noise}). We further compare the properties of the learned representations using different training methods and show that learning representations that are covariant to rotation also aids in boosting performance (Sec.\ref{sec:rot_cov}). We finally compare the results of the proposed method with the state-of-the-art approaches in Sec.\ref{sec:sota_compare}.

\subsection{Experimental Setup}
\label{exp_setup}

We run our experiments either on a single 32GB V100 GPU, or across two such GPUs unless specified otherwise. We train our models with ResNet-18 \cite{resnet} architecture on CIFAR-$10$ and CIFAR-$100$ \cite{krizhevsky2009learning} dataset and with ResNet-50 \cite{resnet} architecture on ImageNet-1k \cite{imagenet_cvpr09} dataset. Our primary evaluations are run for $200$ epochs on CIFAR-$10$ and CIFAR-$100$, and $100$ epochs on ImageNet-100 \cite{imagenet_cvpr09} dataset. We show additional evaluations across varying number of training epochs in Sec.\ref{sec:sota_compare}. We describe the training hyperparameters in Sec.\ref{sec:hyp}. We use the respective base algorithm or the proposed approach to learn the base encoder $f_\theta$, and evaluate its effectiveness by training a linear classifier over this, as is common in prior works \cite{simclr,simsiam,byol,swav}. In this step, the weights of the base encoder are frozen. We additionally report results in a semi-supervised (Sec.\ref{sec:sota_compare}) and transfer learning setting (Sec.\ref{sec:transfer}) as well.

\begin{figure}
     \centering
     \hfill{}
     \begin{subfigure}[b]{0.45\textwidth}
         \centering
         \includegraphics[width=\textwidth]{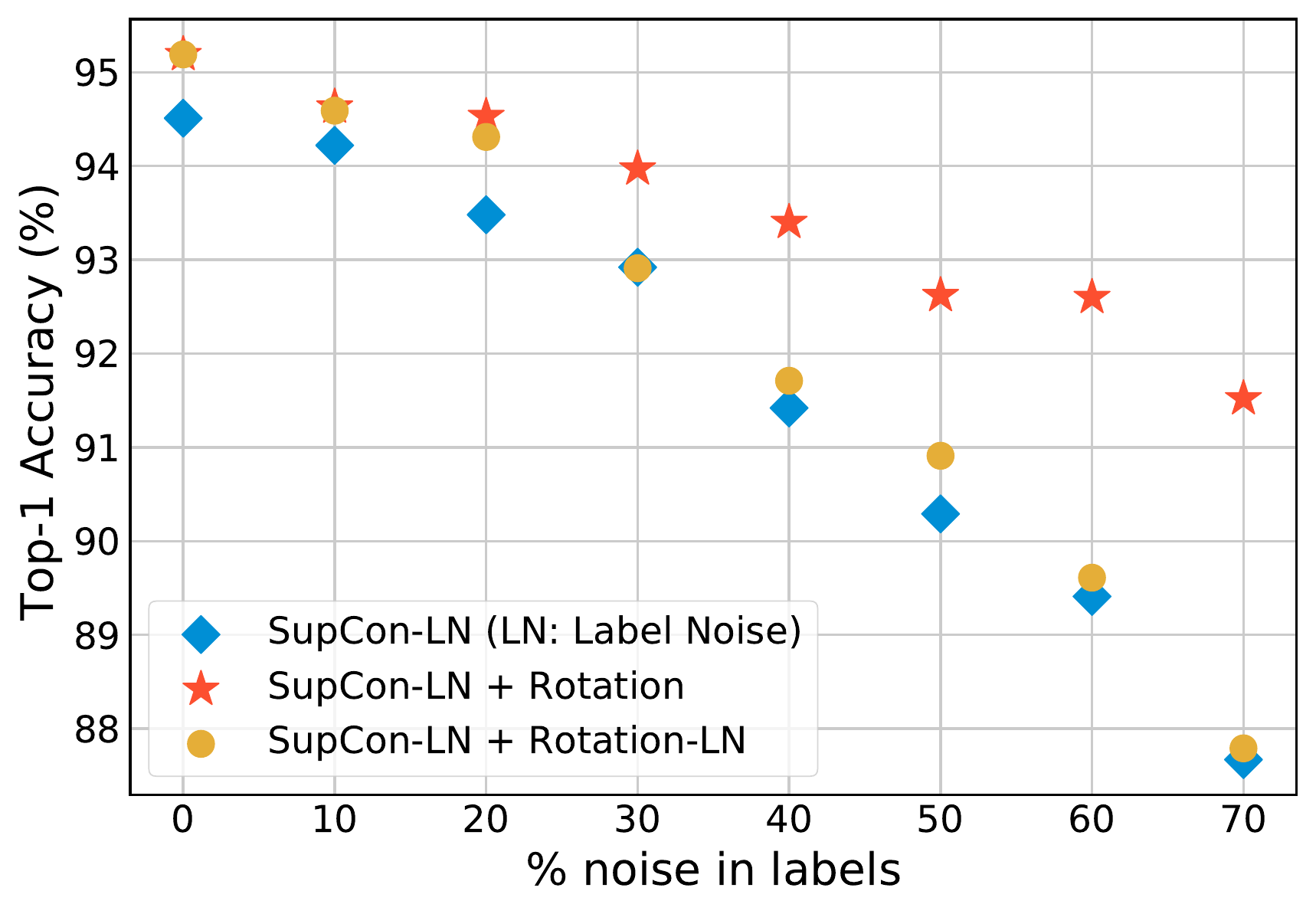}
         \caption{\small{SupCon \citep{khosla2020supervised}}}
         \label{fig:supcon_noise}
     \end{subfigure}
     \hfill
     \begin{subfigure}[b]{0.45\textwidth}
         \centering
         \includegraphics[width=\textwidth]{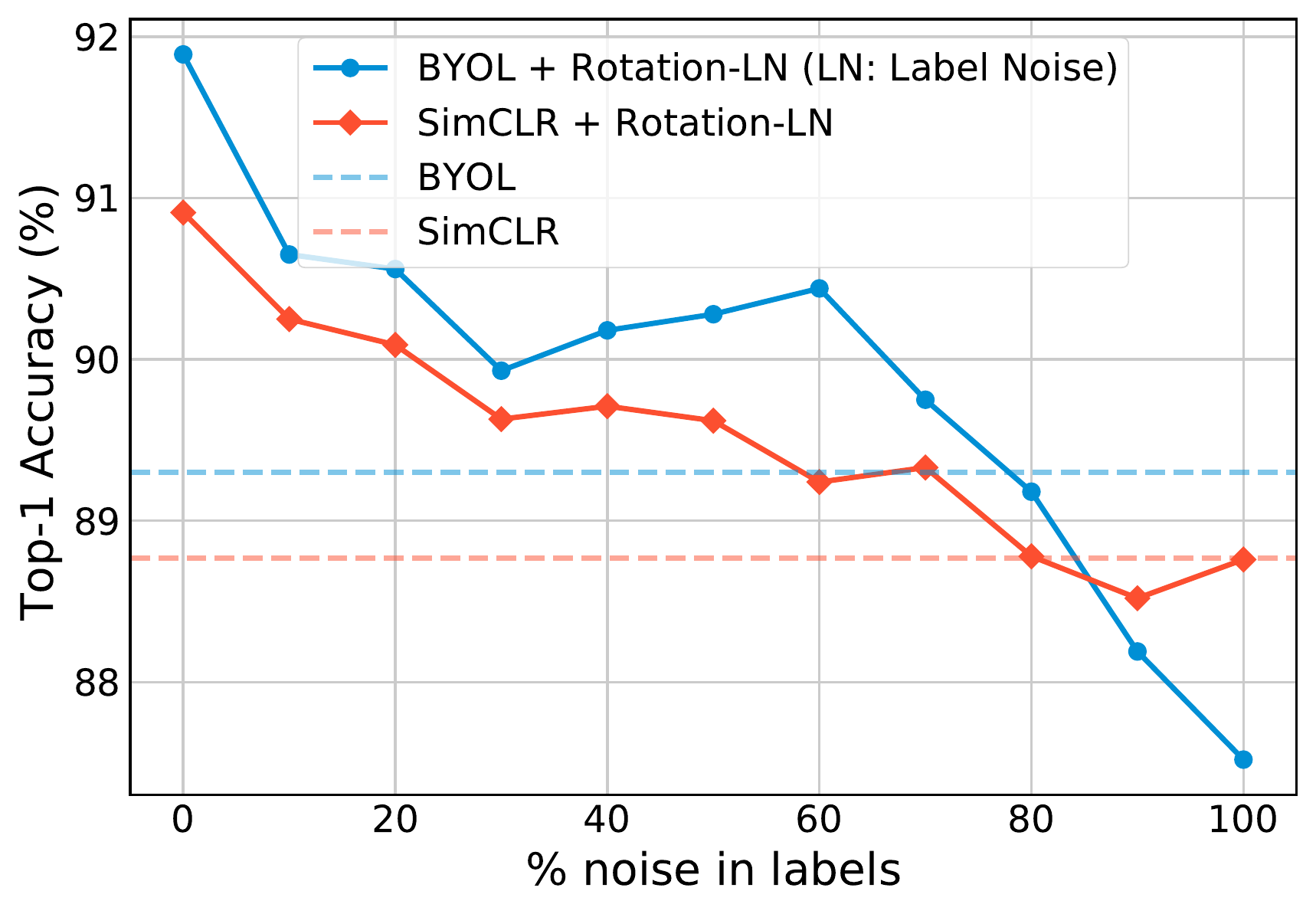}
         \caption{\small{BYOL \citep{byol}, SimCLR \cite{simclr}}}
         \label{fig:byol_noise}
     \end{subfigure}
     \hfill{}

        \caption{\small{The plots demonstrate the impact of label noise in different training objectives on CIFAR-10 dataset. The proposed method (+ Rotation) results in higher performance boost when the amount of label noise in the base method is larger. Addition of label noise to the rotation task reduces the gain in performance.}}
        \label{fig:noise_plots}
\end{figure}

\subsection{Robustness to Noise in the Training objective}
\label{sec:rob_noise}

As discussed in Section-\ref{issues}, instance-similarity based tasks such as SimCLR \cite{simclr} and BYOL \cite{byol} suffer from noise in the training objective, and eliminating this noise can lead to significant performance gains in a fixed training budget. We additionally report results of the proposed approach integrated with SimCLR and BYOL in Tables-\ref{tab:FN_simclr} and \ref{tab:FP_byol} respectively, and obtain gains over the base approach across varying settings of supervision levels. However, as can be seen from the column Gain ($\%$), the gains using the proposed approach reduce with increasing levels of supervision. This is aligned with our hypothesis that the rotation task helps in overcoming the impact of noise in the base instance-similarity task, and therefore, when additional supervision already achieves this objective, gains using the proposed approach are lower.

$\textbf{Label Noise in a Supervised Learning setting:}$ We consider the task of supervised learning using the supervised contrastive (SupCon) learning objective proposed by Khosla et al. \cite{khosla2020supervised}. The training objective is similar to that of SimCLR \cite{simclr} with the exception that same-class negatives are treated as positives. The authors demonstrate that this method outperforms standard supervised training as well. We choose this training objective as this is similar to the instance-similarity based tasks we consider in this paper, while also having significantly lesser noise due to the elimination of false negatives in training. As shown in Fig.\ref{fig:supcon_noise}, even in this setting, the proposed method achieves $0.68\%$ improvement, achieving a new state-of-the-art in supervised learning. In order to highlight the impact of noise in training, we run a controlled set of experiments by adding a fixed amount of label noise in each run. The plot in Fig. \ref{fig:supcon_noise} shows the trend in accuracy of the SupCon algorithm with increasing label noise. The proposed method achieves a significant boost over the SupCon baseline consistently across different noise levels. Further, as the amount of noise in training increases, we achieve higher gains using the proposed approach, indicating that the rotation task is indeed helping overcome noise in the training objective.

We also consider a set of experiments where an equal amount of label noise is added to the SupCon training objective and to the rotation prediction task. We note that in majority of the runs (excluding the case of noise above $70\%$), the accuracy is very similar to the SupCon baseline with the same amount of noise. This indicates that the knowledge of true labels in handcrafted tasks such as rotation prediction is the key factor that contributes to the improvement achieved using the proposed approach. 

We perform the experiments of adding label noise to the rotation prediction task when combined with BYOL and SimCLR as well. As shown in Fig.\ref{fig:byol_noise} we find that the gains with the rotation prediction task drops considerably over $0-20\%$ label noise, indicating that a similar amount of noise ($\sim20\%$) is present in the BYOL/ SimCLR training objectives as well. Further, addition of rotation prediction task helps marginally ($0.47-1.38\%$) even with higher amount of noise ($30-60\%$) in rotation annotations. This indicates that, while the rotation prediction primarily helps by providing a noise-free training objective, it aids the main task in other ways too. We investigate this in the following section.

\subsection{Learning rotation-covariant representations}
\label{sec:rot_cov}

The task of enforcing similarity across various augmentations of a given image yields representations that are invariant to such transformations. In sharp contrast, the representations learned by humans are covariant with respect to factors such as rotation, color and scale, although we are able to still correlate multiple transformations of the same object very well. This hints at the fact that learning covariant representations could help the accuracy of downstream tasks such as object detection and classification.

In Table-\ref{tab:eval_no_noise}, we compare the rotation sensitivity and contrastive task accuracy of representations at the output of the base encoder $f_\theta$, and the projection network $g_\theta$. We follow the process described by Chen et al. \cite{simclr} to obtain these results. We freeze the network till the respective layer ($f_\theta$ or $g_\theta$) and train a rotation task classifier over this using a $2$-layer MLP head. We measure the rotation task accuracy, which serves as an indication of the amount of rotation sensitivity in the base network. We further compute the contrastive task accuracy on the representations learned, by checking whether the two augmentations of a given image are more similar to each other when compared to augmentations of other images in the same batch. 

\begin{table}
\begin{minipage}{0.48\linewidth}
\caption{\textbf{Task Performance ($\%$):} Evaluation of representations learned using various algorithms on the task of rotation prediction and instance-discrimination.}
\small
\centering
\label{tab:eval_no_noise}
\resizebox{1.0\linewidth}{!}{
\begin{tabular}{lccccc}
\toprule
Method & \multirow{2}{*}{Linear} & \multicolumn{2}{c}{Rotation Acc} & \multicolumn{2}{c}{Contrastive Acc} \\
 &  & f(.) & g(f(.)) & f(.) & g(f(.)) \\
 \midrule
Supervised & 94.03 & 80.54 & - & 46.36 & - \\
BYOL & 89.30 & 73.40 & 58.32 & 78.53 & 78.82 \\
Rotation & 84.00 & 93.69 & 93.46 & 31.61 & 1.52 \\
BYOL+Rotation & 91.89 & 93.73 & 93.54 & 72.85 & 67.81 \\
\bottomrule
\end{tabular}
}
\end{minipage}
\hfill
\begin{minipage}{0.48\linewidth}
\caption{BYOL + Rotation with \textbf{varying noise in the rotation labels}. Rotation prediction accuracy correlates with linear evaluation accuracy.}

\small
\centering
\label{tab:with_noise}
\resizebox{1.0\linewidth}{!}{
\begin{tabular}{lcccccc}
\toprule
 Rotation & \multirow{2}{*}{Linear} &  \multicolumn{2}{c}{Rotation Acc} & \multicolumn{2}{c}{Contrastive Acc} \\
  Noise & & f(.) & g(f(.) & f(.) & g(f(.) \\
 \midrule
$30\%$ & 89.93 & 91.88 & 91.78 & 73.42 & 64.25 \\
$50\%$ & 90.28 & 89.95 & 85.82 & 78.18 & 77.39 \\
$70\%$ & 89.75 & 80.49 & 67.26 & 78.55 & 77.31 \\
$80\%$ & 89.18 & 77.43 & 63.53 & 77.26 & 76.92 \\
\bottomrule
\end{tabular}
}
\end{minipage}

\end{table}

Interestingly, a fully supervised network is more sensitive to rotation ($80.54\%$) when compared to the representations learned using BYOL ($73.4\%$). Chen et al. \cite{simclr} also show that rotation augmentation hurts performance of SimCLR. These observations indicate that invariance to rotation hurts performance, and reducing this lead to better representations. While RotNet has higher accuracy on the rotation task, it does significantly worse on the instance discrimination task, leading to sub-optimal performance compared to BYOL. In the proposed method, we achieve better rotation task accuracy with a small drop in the contrastive task accuracy when compared to BYOL. This also results in an overall higher performance after Linear evaluation.

We also investigate rotation invariance for the experiments in Sec.\ref{sec:rob_noise} with BYOL as the base method, where noise is added to the rotation task. As shown in Table-\ref{tab:with_noise}, we find that as the amount of noise increases in the rotation task, the amount of rotation invariance increases, leading to a drop in accuracy. Even with $50\%$ noise in the rotation task, we achieve $16.55\%$ boost in rotation performance, leading to $0.98\%$ improvement in the accuracy after linear evaluation. Since the BYOL learning task possibly contains lesser noise compared to this, the gain in performance can be justified by the fact that rotation-covariant representations lead to improved performance on natural image datasets.

\subsection{Comparison with the state-of-the-art}

\label{sec:sota_compare}

We compare the performance of the proposed method with the respective baselines in the setting of linear evaluation on CIFAR-10, CIFAR-100 (Table-\ref{table:cifar}), ImageNet-100 and ImageNet-1k (Table-\ref{tab:imagenet2}) datasets. We perform extensive hyperparameter search to obtain reliable results on the baseline methods for CIFAR-10 and CIFAR-100, since most existing works report the optimal settings for ImageNet-1k training alone. As shown in Table-\ref{table:cifar}, although the performance of Rotation prediction \cite{rotnet} itself is significantly worse that other methods, we obtain gains of $2.14\%$, $2.59\%$, $3.6\%$ and $2.14\%$ on CIFAR-10 and $2.44\%$, $6.4\%$, $7.1\%$ and $3.11\%$ on CIFAR-100 by using the proposed method with SimCLR \cite{simclr}, BYOL \cite{byol}, SwAV \cite{swav} and SimSiam \cite{simsiam} respectively. 

\begin{table}
\caption{\textbf{CIFAR-10, CIFAR-100:} Accuracy ($\%$) of the proposed method compared to baselines under two evaluation settings - K-Nearest Neighbor (KNN) classification with K=200 and Linear classifier training. The proposed method achieves significant performance gains across all settings.}
\centering
% \resizebox{0.9\linewidth}{!}{
\label{table:cifar}
\begin{tabular}{ccccc}
\toprule
 & \multicolumn{2}{c}{CIFAR-10 (200 epochs)} & \multicolumn{2}{c}{CIFAR-100 (200 epochs)} \\
Method & KNN & Linear &  KNN & Linear  \\
\midrule
Rotation Pred. \cite{rotnet}~~~~ & 78.01 & 84.00  & 36.25 & 50.87  \\
SimCLR \cite{simclr}& 86.37 & 88.77 & 55.10 & 62.96  \\
SimCLR + Ours & 88.69 & 90.91  & 57.09 & 65.40 \\
BYOL \cite{byol} & 86.56 & 89.30  & 54.37 & 60.67  \\
BYOL + Ours & 89.80 & 91.89  & 58.41 & 67.03  \\

SwAV \cite{swav}&	80.65 &	83.60	& 40.35&	51.50 \\
SwAV + Ours &	85.26&	87.20&	50.09&	58.60\\
SimSiam	\cite{simsiam}&87.05&	89.77&	56.90&	64.27\\
SimSiam + Ours&	\textbf{90.35}&	\textbf{91.91}	&\textbf{58.92}	&\textbf{67.38}\\

\bottomrule
\end{tabular}
% }
\end{table}

\begin{figure}[h!]
     \centering
     \hfill{}
     \begin{subfigure}[b]{0.4\textwidth}
         \centering
         \includegraphics[width=\textwidth]{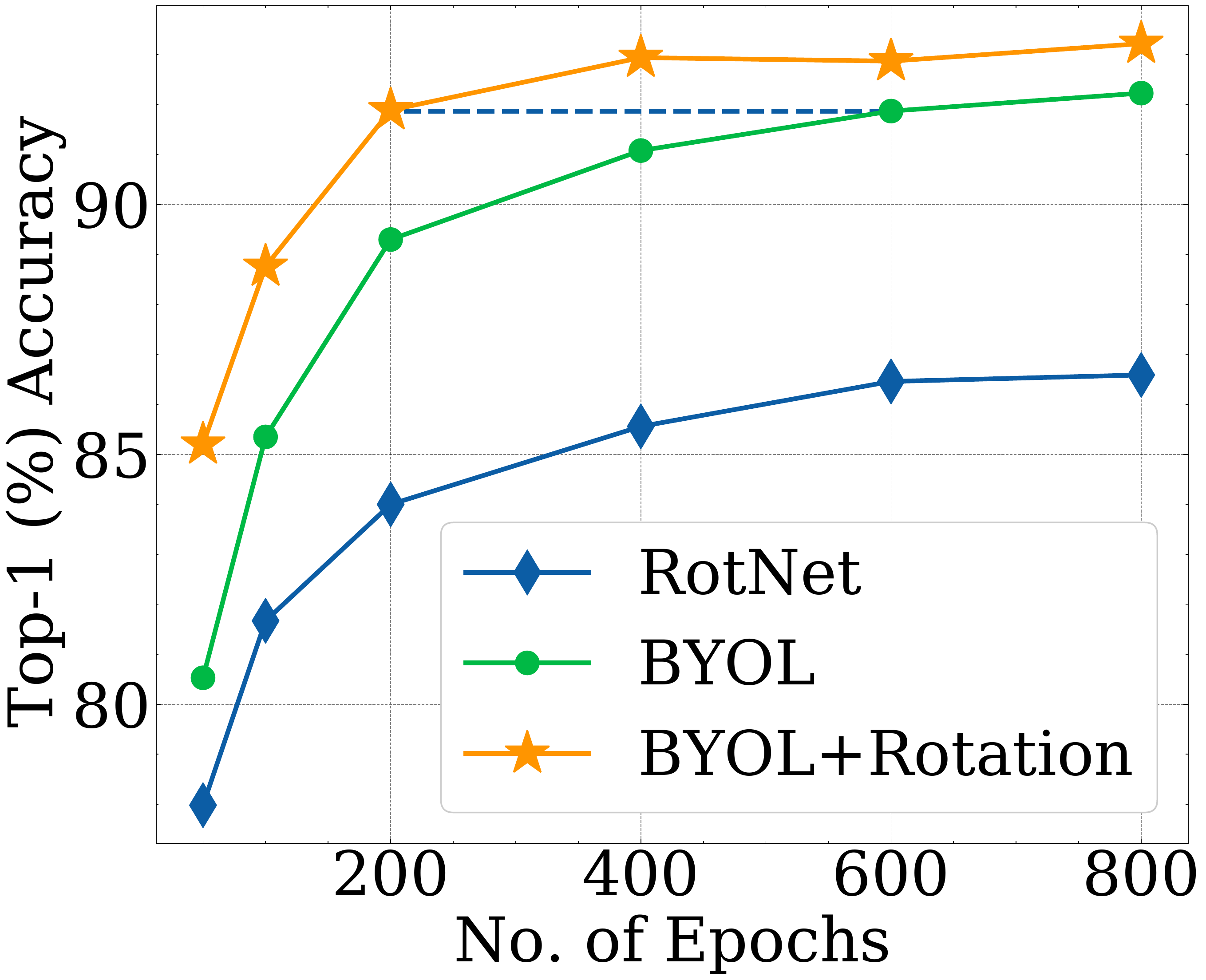}
         \caption{\small{Top-1 Accuracy}}
         \label{fig:acc_cifar_conv}
     \end{subfigure}
     \hfill
     \begin{subfigure}[b]{0.4\textwidth}
         \centering
         \includegraphics[width=\textwidth]{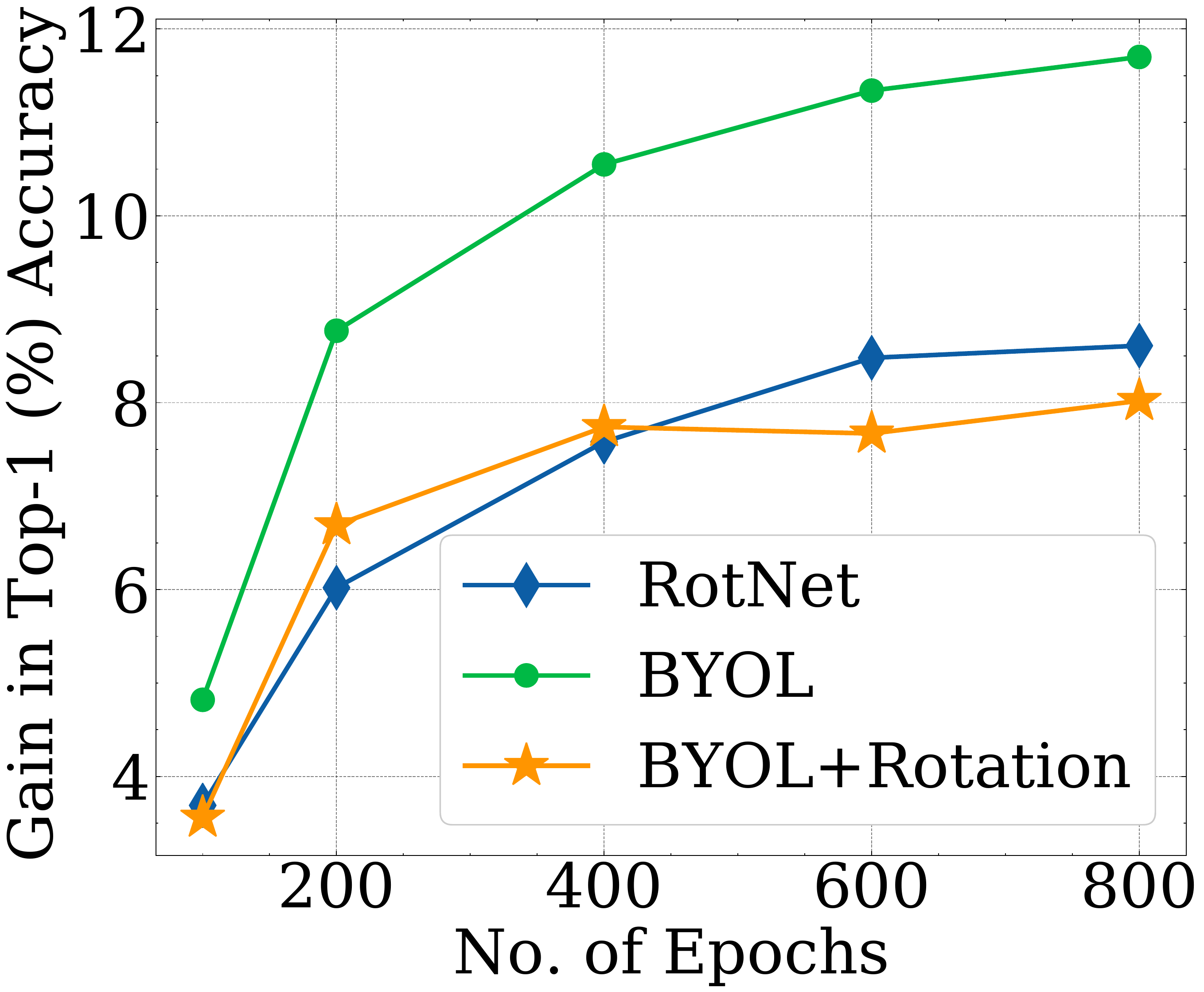}
         \caption{\small{Gain in Top-1 Accuracy}}
         \label{fig:acc_cifar_gain}
     \end{subfigure}
     \hfill{}
        \caption{\small{(a) Accuracy ($\%$) after Linear layer training for BYOL \cite{byol}, RotNet \cite{rotnet} and the proposed method (BYOL+Rotation) on CIFAR-10. The proposed method achieves the same accuracy as the baseline in one-third the training time (shown using blue dotted line). (b) Gain in Top-1 Accuracy ($\%$), or the difference between accuracy of the current epoch and epoch-50. 
        Plot (a) shows the improvement in effectiveness of the proposed approach and plot (b) shows the improvement in efficiency or convergence rate.}}
        \label{fig:acc_cifar_all}
\end{figure}

\begin{table}[h]
\caption{\textbf{ImageNet-100 and ImageNet-1k:} Performance ($\%$) of the proposed method when compared to baselines under three evaluation settings - Linear classifier training and Semi-Supervised Learning with $1\%$ and $10\%$ labels. The proposed method achieves significant performance gains.}
\centering
\label{tab:imagenet2}
% \resizebox{0.9\linewidth}{!}{
\begin{tabular}{lcccccc}
\toprule
\multirow{2}{*}{Method} & Linear Acc & \multicolumn{2}{c}{\begin{tabular}[c]{@{}c@{}}Semi-Supervised\\ 1$\%$ labels\end{tabular}} & \multicolumn{2}{c}{\begin{tabular}[c]{@{}c@{}}Semi-Supervised\\ 10$\%$ labels\end{tabular}} \\
                        & Top-1       & Top-1                                      & Top-5                                     & Top-1                                      & Top-5          \\
\midrule
\multicolumn{5}{c}{\textbf{ImageNet-100 (100 epochs, ResNet-18)}}                                                                                                                                                            \\
\midrule
Rotation Prediction  \cite{rotnet}                & 53.86       & 34.72                                      & 65.70                                     & 51.18                                      & 81.38                                      \\
BYOL   \cite{byol}                 & 71.02       & 46.60                                      & 75.50                                     & 68.00                                      & 89.80                                      \\
BYOL + Ours             & 73.60       & 56.40                                      & 83.50                                     & 72.30                                      & 91.40                                      \\
SimCLR \cite{simclr}                 & 72.02       & 57.28                                      & 83.69                                     & 71.44                                      & 91.72                                      \\
SimCLR + Ours           & 73.24       & \textbf{57.80}                                      & \textbf{83.84}                                     & \textbf{72.52}                                      & \textbf{92.10}                                      \\
SwAV \cite{swav}                   & 72.20       & 49.38                                      & 78.41                                     & 67.56                                      & 90.78                                      \\
SwAV + Ours             & \textbf{74.40}       & 52.02                                      & 80.01                                     & 69.68                                      & 91.43                                      \\
\midrule
\multicolumn{5}{c}{\textbf{ImageNet-1k (30 epochs, ResNet-50)}}                                                                                                                                                                         \\
\midrule
SwAV   \cite{swav}                 & 54.90       & 32.20                                      & 58.20                                     & 51.82                                      & 77.60                                      \\
SwAV + Ours             & \textbf{57.30}       & \textbf{32.80}                                      & \textbf{59.12}                                     & \textbf{53.80}                                      & \textbf{78.54}            \\
\bottomrule
\end{tabular}
\end{table}

We present results on CIFAR-10 dataset with varying number of training epochs in Fig.\ref{fig:acc_cifar_conv} using BYOL as the base approach. Across all settings, we obtain improved results over the BYOL baseline. The  proposed  method achieves the same accuracy as the baseline in one-third the training time (shown using blue dotted line) as shown in Fig.\ref{fig:acc_cifar_conv}. We show the difference in accuracy with respect to accuracy obtained with 50 epochs of training in Fig.\ref{fig:acc_cifar_gain}, to clearly visualize the convergence rate of different methods. It can be seen that the proposed method has a similar convergence trend as the Rotation task, while outperforming BYOL in terms of Top-1 Accuracy, highlighting that integrating these methods indeed combines the benefits of both methods.

We present results on ImageNet-100 dataset in Table-\ref{tab:imagenet2}. To limit the computational cost on our ImageNet-100 and ImageNet-1k runs, we either use the tuned hyperparameters from the official repository, or follow the settings from other popular repositories that report competent results. Due to the unavailability of tuned hyperparameters on this dataset for SimSiam, we skip reporting results of this method on ImageNet-100. We achieve gains of $2.58\%$, $1.22\%$ and $2.2\%$ on BYOL \cite{byol}, SimCLR \cite{simclr} and SwAV \cite{swav} respectively in Top-1 accuracy. We obtain the best results by integrating the proposed method with SwAV, and hence report ImageNet-1k results on the same method, in order to demonstrate the scalability of the proposed method to a large-scale dataset. We present the result of $30$-epochs of training on ImageNet-1k in Table-\ref{tab:imagenet2}. Using the proposed approach, we obtain a boost of $2.4\%$ in Top-$1$ accuracy over the SwAV baseline. We present additional results on longer training epochs in Sec.\ref{imagenet_supp}.

Furthermore, we present results on ImageNet-100 dataset with varying number of training epochs in Fig.\ref{fig:acc_across_epochs}. Using the proposed method, we achieve gains across all settings with respect to the number of training epochs. 
We obtain improved results over the base methods in semi-supervised learning (Table-\ref{tab:imagenet2}) and transfer learning settings as well. We discuss the transfer learning results in Sec.\ref{sec:transfer}. 

\begin{table}[h]

\begin{minipage}{0.4\linewidth}

\caption{\small{\textbf{Combining BYOL with handcrafted pretext tasks:} Accuracy in ($\%$) after linear evaluation, of various algorithms on ImageNet-100 dataset.
}}

\resizebox{1.0\linewidth}{!}{
\label{table:other_tasks}
\centering
\begin{tabular}{lcc}
\toprule
                         & \textbf{Top-1 ($\%$)} & \textbf{Top-5 ($\%$)} \\
                         \midrule
RotNet (R)                 & 53.86          & 81.26                         \\
Jigsaw (J)                  & 42.01          & 72.10                         \\
BYOL                    & 71.02          & 91.78                         \\
BYOL + R          & 73.60          & \textbf{92.98}                \\
BYOL + J            & 73.60          & 92.72                         \\
BYOL + J + R & \textbf{74.72} & 92.94 \\
\bottomrule
\end{tabular}}
\end{minipage}
\hfill
\begin{minipage}{0.55\linewidth}
\centering
          \includegraphics[width=\linewidth]{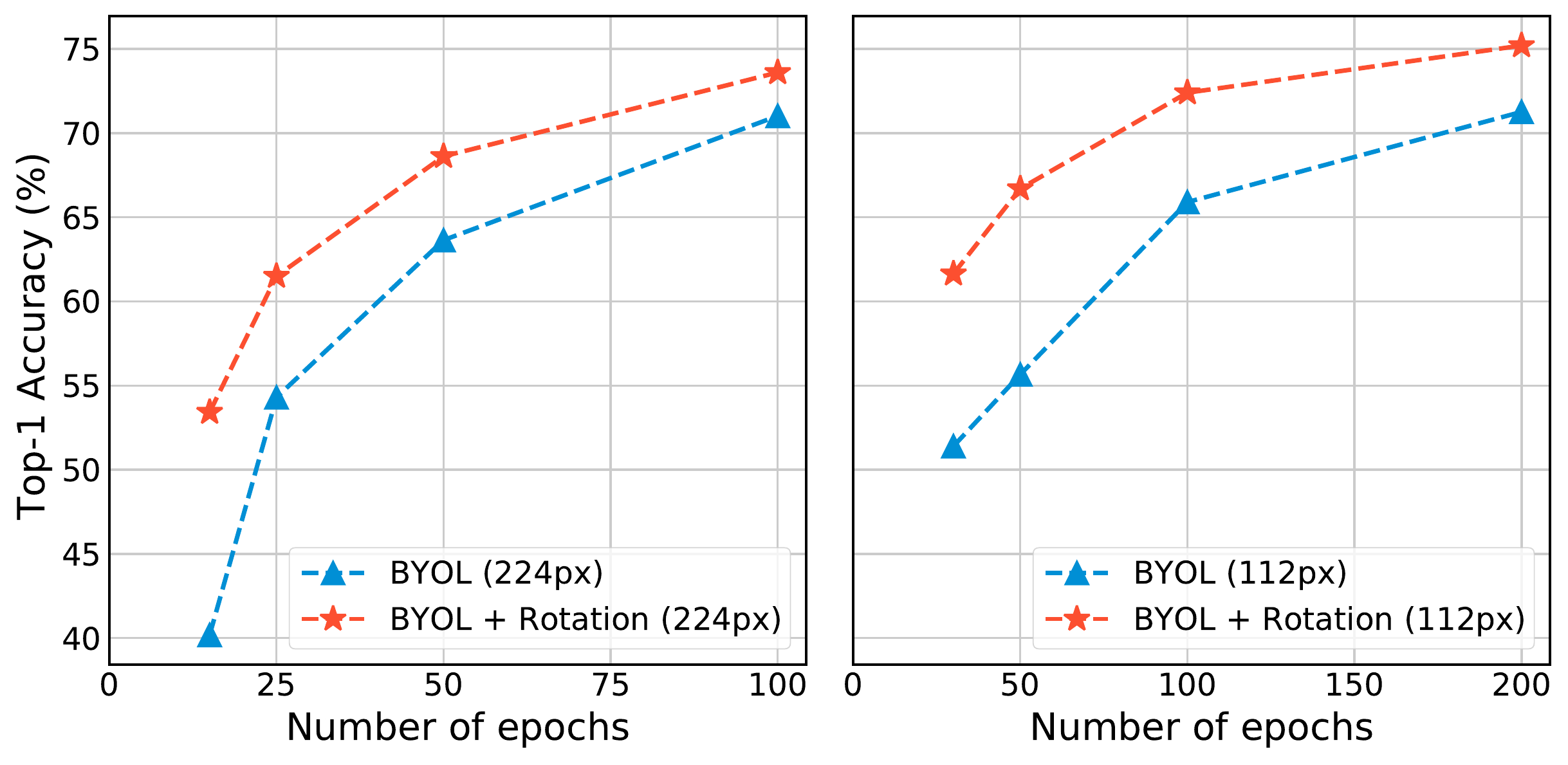}
         \label{fig:byol_in100_epochs}
         
        \captionof{figure}{\small{Accuracy ($\%$) after Linear layer training for BYOL and the proposed method (BYOL+Rotation) for ImageNet-100. The proposed method achieves significant gains over the baseline in all settings.}}
        \label{fig:acc_across_epochs}
        
\end{minipage}

\end{table}

\subsection{Integration with other tasks}

In this work, we empirically show that combining instance-discriminative tasks with well-posed handcrafted pretext tasks such as Rotation prediction \cite{rotnet} can indeed lead to more effective and efficient learning of visual representations. While we choose the Rotation prediction task due to its simplicity in implementation, and applicability to low resolution images (such as CIFAR-10), it is indeed possible to achieve gains by using other well-posed tasks as well. In Table-\ref{table:other_tasks}, we report results on the ImageNet-100 \cite{cmc} dataset by combining the base BYOL \cite{byol} algorithm individually with Rotation prediction \cite{rotnet}, Jigsaw puzzle solving \cite{jigsaw} and both. Although the Jigsaw puzzle solving task is sub-optimal when compared to the Rotation prediction task, we achieve similar gains in performance when these tasks are combined with BYOL. We obtain the best gains ($3.7\%$) when we combine both tasks with BYOL. This shows that the analysis on well-defined tasks being able to aid the learning of instance-discriminative tasks that are noisy is indeed generic, and not specific to the Rotation prediction task alone.

\section{Conclusions}
In this work, we investigate reasons for the slow convergence of recent instance-similarity based methods, and propose to improve the same by jointly training them with well-posed tasks such as rotation prediction. While instance-discriminative approaches learn better representations, handcrafted tasks have the advantage of faster convergence as the training objective is well defined and there is typically no (or very less) noise in the generated pseudo-labels. The complementary nature of the two kinds of tasks makes it suitable to achieve the gains associated with both by combining them. Using the proposed approach, we show significant gains in performance under a fixed training budget, along with improvements in training efficiency. We show similar gains in performance by combining the base algorithms with the task of Jigsaw puzzle solving as well. We hope that our work will revive research interest in designing specialized tasks, so that they can be help boost the effectiveness and efficiency of state-of-the-art methods. 

\section{Acknowledgments}
This work was supported by the Qualcomm Innovation Fellowship. We are thankful for the support.

\bibliographystyle{splncs04}
\bibliography{references}

\begin{thebibliography}{10}
\providecommand{\url}[1]{\texttt{#1}}
\providecommand{\urlprefix}{URL }
\providecommand{\doi}[1]{https://doi.org/#1}

\bibitem{asano2019self}
Asano, Y.M., Rupprecht, C., Vedaldi, A.: Self-labelling via simultaneous
  clustering and representation learning. In: International Conference on
  Learning Representations (ICLR) (2020)

\bibitem{bossard2014food}
Bossard, L., Guillaumin, M., Van~Gool, L.: Food-101--mining discriminative
  components with random forests. In: European conference on computer vision.
  pp. 446--461. Springer (2014)

\bibitem{caron2018deep}
Caron, M., Bojanowski, P., Joulin, A., Douze, M.: Deep clustering for
  unsupervised learning of visual features. In: Proceedings of the European
  Conference on Computer Vision (ECCV) (2018)

\bibitem{swav}
Caron, M., Misra, I., Mairal, J., Goyal, P., Bojanowski, P., Joulin, A.:
  Unsupervised learning of visual features by contrasting cluster assignments.
  In: Advances in Neural Information Processing Systems (NeurIPS) (2020)

\bibitem{simclr}
Chen, T., Kornblith, S., Norouzi, M., Hinton, G.: A simple framework for
  contrastive learning of visual representations. In: Proceedings of the 37th
  International Conference on Machine Learning (ICML) (2020)

\bibitem{chen2019self}
Chen, T., Zhai, X., Ritter, M., Lucic, M., Houlsby, N.: Self-supervised gans
  via auxiliary rotation loss. In: Proceedings of the IEEE/CVF Conference on
  Computer Vision and Pattern Recognition (CVPR) (2019)

\bibitem{simsiam}
Chen, X., He, K.: Exploring simple siamese representation learning. In:
  Proceedings of the IEEE/CVF Conference on Computer Vision and Pattern
  Recognition (CVPR) (2021)

\bibitem{cimpoi2014describing}
Cimpoi, M., Maji, S., Kokkinos, I., Mohamed, S., Vedaldi, A.: Describing
  textures in the wild. In: Proceedings of the IEEE Conference on Computer
  Vision and Pattern Recognition. pp. 3606--3613 (2014)

\bibitem{turrisi2021sololearn}
da~Costa, V.G.T., Fini, E., Nabi, M., Sebe, N., Ricci, E.: Solo-learn: A
  library of self-supervised methods for visual representation learning (2021),
  \url{https://github.com/vturrisi/solo-learn}

\bibitem{cuturi2013sinkhorn}
Cuturi, M.: Sinkhorn distances: Lightspeed computation of optimal transport.
  In: Advances in Neural Information Processing Systems (NeurIPS) (2013)

\bibitem{dangovski2021equivariant}
Dangovski, R., Jing, L., Loh, C., Han, S., Srivastava, A., Cheung, B., Agrawal,
  P., Soljacic, M.: Equivariant self-supervised learning: Encouraging
  equivariance in representations. In: International Conference on Learning
  Representations (ICLR) (2022)

\bibitem{imagenet_cvpr09}
Deng, J., Dong, W., Socher, R., Li, L.J., Li, K., Fei-Fei, L.: {ImageNet: A
  Large-Scale Hierarchical Image Database}. In: Proceedings of the IEEE
  Conference on Computer Vision and Pattern Recognition (CVPR) (2009)

\bibitem{context}
Doersch, C., Gupta, A., Efros, A.A.: Unsupervised visual representation
  learning by context prediction. In: IEEE International Conference on Computer
  Vision (ICCV) (2015)

\bibitem{doersch2017multi}
Doersch, C., Zisserman, A.: Multi-task self-supervised visual learning. In:
  IEEE International Conference on Computer Vision (ICCV) (2017)

\bibitem{donahue2016adversarial}
Donahue, J., Kr{\"a}henb{\"u}hl, P., Darrell, T.: Adversarial feature learning.
  In: International Conference on Learning Representations (ICLR) (2017)

\bibitem{ericsson2021well}
Ericsson, L., Gouk, H., Hospedales, T.M.: How well do self-supervised models
  transfer? In: Proceedings of the IEEE/CVF Conference on Computer Vision and
  Pattern Recognition. pp. 5414--5423 (2021)

\bibitem{everingham2010pascal}
Everingham, M., Van~Gool, L., Williams, C.K., Winn, J., Zisserman, A.: The
  pascal visual object classes (voc) challenge. International journal of
  computer vision  \textbf{88}(2),  303--338 (2010)

\bibitem{faster2015towards}
Faster, R.: Towards real-time object detection with region proposal networks.
  Advances in neural information processing systems  \textbf{9199} (2015)

\bibitem{fei2004learning}
Fei-Fei, L., Fergus, R., Perona, P.: Learning generative visual models from few
  training examples: An incremental bayesian approach tested on 101 object
  categories. In: 2004 conference on computer vision and pattern recognition
  workshop. pp. 178--178. IEEE (2004)

\bibitem{rotnet}
Gidaris, S., Singh, P., Komodakis, N.: Unsupervised representation learning by
  predicting image rotations. In: International Conference on Learning
  Representations (ICLR) (2018)

\bibitem{goodfellow2014generative}
Goodfellow, I., Pouget-Abadie, J., Mirza, M., Xu, B., Warde-Farley, D., Ozair,
  S., Courville, A., Bengio, Y.: Generative adversarial nets. In: Advances in
  Neural Information Processing Systems (NeurIPS) (2014)

\bibitem{byol}
Grill, J.B., Strub, F., Altch\'{e}, F., Tallec, C., Richemond, P., Buchatskaya,
  E., Doersch, C., Avila~Pires, B., Guo, Z., Gheshlaghi~Azar, M., Piot, B.,
  kavukcuoglu, k., Munos, R., Valko, M.: Bootstrap your own latent - a new
  approach to self-supervised learning. In: Advances in Neural Information
  Processing Systems (NeurIPS) (2020)

\bibitem{moco}
He, K., Fan, H., Wu, Y., Xie, S., Girshick, R.: Momentum contrast for
  unsupervised visual representation learning. In: Proceedings of the IEEE/CVF
  Conference on Computer Vision and Pattern Recognition (CVPR) (2020)

\bibitem{resnet}
He, K., Zhang, X., Ren, S., Sun, J.: Deep residual learning for image
  recognition. In: IEEE Conference on Computer Vision and Pattern Recognition
  (CVPR) (2016)

\bibitem{datacpc}
H\'{e}naff, O.J., Srinivas, A., De~Fauw, J., Razavi, A., Doersch, C., Eslami,
  S.M.A., Van Den~Oord, A.: Data-efficient image recognition with contrastive
  predictive coding. In: Proceedings of the 37th International Conference on
  Machine Learning (ICML) (2020)

\bibitem{patrick-hua}
Hua, T.: A pytorch implementation for paper, exploring simple siamese
  representation learning (2021), \url{https://github.com/PatrickHua/SimSiam},
  https://github.com/PatrickHua/SimSiam

\bibitem{batchnorm}
Ioffe, S., Szegedy, C.: Batch normalization: Accelerating deep network training
  by reducing internal covariate shift. In: International conference on machine
  learning. pp. 448--456. PMLR (2015)

\bibitem{khosla2020supervised}
Khosla, P., Teterwak, P., Wang, C., Sarna, A., Tian, Y., Isola, P., Maschinot,
  A., Liu, C., Krishnan, D.: Supervised contrastive learning. In: Advances in
  Neural Information Processing Systems (NeurIPS) (2020)

\bibitem{kinakh2021scatsimclr}
Kinakh, V., Voloshynovskiy, S., Taran, O.: Scatsim{CLR}: self-supervised
  contrastive learning with pretext task regularization for small-scale
  datasets. In: 2nd Visual Inductive Priors for Data-Efficient Deep Learning
  Workshop (2021)

\bibitem{kingma2013auto}
Kingma, D.P., Welling, M.: Auto-encoding variational bayes. arXiv preprint
  arXiv:1312.6114  (2013)

\bibitem{kornblith2019better}
Kornblith, S., Shlens, J., Le, Q.V.: Do better imagenet models transfer better?
  In: Proceedings of the IEEE/CVF Conference on Computer Vision and Pattern
  Recognition. pp. 2661--2671 (2019)

\bibitem{krause2013collecting}
Krause, J., Deng, J., Stark, M., Fei-Fei, L.: Collecting a large-scale dataset
  of fine-grained cars  (2013)

\bibitem{alexnet}
Krizhevsky, A., Sutskever, I., Hinton, G.E.: Imagenet classification with deep
  convolutional neural networks. In: Advances in Neural Information Processing
  Systems (NeurIPS) (2012)

\bibitem{krizhevsky2009learning}
Krizhevsky, A., et~al.: Learning multiple layers of features from tiny images
  (2009)

\bibitem{lecun2015deep}
LeCun, Y., Bengio, Y., Hinton, G.: Deep learning. nature  \textbf{521}(7553),
  436--444 (2015)

\bibitem{pcl}
Li, J., Zhou, P., Xiong, C., Socher, R., Hoi, S.C.: Prototypical contrastive
  learning of unsupervised representations. arXiv preprint arXiv:2005.04966
  (2020)

\bibitem{lin2017feature}
Lin, T.Y., Doll{\'a}r, P., Girshick, R., He, K., Hariharan, B., Belongie, S.:
  Feature pyramid networks for object detection. In: Proceedings of the IEEE
  conference on computer vision and pattern recognition. pp. 2117--2125 (2017)

\bibitem{maji2013fine}
Maji, S., Rahtu, E., Kannala, J., Blaschko, M., Vedaldi, A.: Fine-grained
  visual classification of aircraft. arXiv preprint arXiv:1306.5151  (2013)

\bibitem{pirl}
Misra, I., Maaten, L.v.d.: Self-supervised learning of pretext-invariant
  representations. In: Proceedings of the IEEE/CVF Conference on Computer
  Vision and Pattern Recognition (CVPR) (2020)

\bibitem{mitrovic2020icbinb}
Mitrovic, J., McWilliams, B., Rey, M.: Less can be more in contrastive
  learning. In: "I Can't Believe It's Not Better!" NeurIPS Workshop (2020)

\bibitem{nilsback2008automated}
Nilsback, M.E., Zisserman, A.: Automated flower classification over a large
  number of classes. In: 2008 Sixth Indian Conference on Computer Vision,
  Graphics \& Image Processing. pp. 722--729. IEEE (2008)

\bibitem{jigsaw}
Noroozi, M., Favaro, P.: Unsupervised learning of visual representations by
  solving jigsaw puzzles. In: European Conference on Computer Vision (ECCV)
  (2016)

\bibitem{counting}
Noroozi, M., Pirsiavash, H., Favaro, P.: Representation learning by learning to
  count. In: Proceedings of the IEEE International Conference on Computer
  Vision (ICCV) (2017)

\bibitem{cpc}
Oord, A.v.d., Li, Y., Vinyals, O.: Representation learning with contrastive
  predictive coding. arXiv preprint arXiv:1807.03748  (2018)

\bibitem{parkhi2012cats}
Parkhi, O.M., Vedaldi, A., Zisserman, A., Jawahar, C.: Cats and dogs. In: 2012
  IEEE conference on computer vision and pattern recognition. pp. 3498--3505.
  IEEE (2012)

\bibitem{inpainting}
Pathak, D., Krahenbuhl, P., Donahue, J., Darrell, T., Efros, A.A.: Context
  encoders: Feature learning by inpainting. In: Proceedings of the IEEE
  conference on computer vision and pattern recognition (CVPR) (2016)

\bibitem{gradcam}
Selvaraju, R.R., Cogswell, M., Das, A., Vedantam, R., Parikh, D., Batra, D.:
  Grad-cam: Visual explanations from deep networks via gradient-based
  localization. In: Proceedings of the IEEE international conference on
  computer vision (ICCV) (2017)

\bibitem{selvaraju2020casting}
Selvaraju, R.R., Desai, K., Johnson, J., Naik, N.: Casting your model: Learning
  to localize improves self-supervised representations. In: Proceedings of the
  IEEE/CVF Conference on Computer Vision and Pattern Recognition (CVPR) (2021)

\bibitem{inception}
Szegedy, C., Liu, W., Jia, Y., Sermanet, P., Reed, S., Anguelov, D., Erhan, D.,
  Vanhoucke, V., Rabinovich, A.: Going deeper with convolutions. In:
  Proceedings of the IEEE conference on computer vision and pattern recognition
  (CVPR) (2015)

\bibitem{cmc}
Tian, Y., Krishnan, D., Isola, P.: Contrastive multiview coding. In: European
  conference on computer vision (ECCV) (2020)

\bibitem{wu2019detectron2}
Wu, Y., Kirillov, A., Massa, F., Lo, W.Y., Girshick, R.: Detectron2.
  https://github.com/facebookresearch/detectron2 (2019)

\bibitem{xiao2010sun}
Xiao, J., Hays, J., Ehinger, K.A., Oliva, A., Torralba, A.: Sun database:
  Large-scale scene recognition from abbey to zoo. In: 2010 IEEE computer
  society conference on computer vision and pattern recognition. pp.
  3485--3492. IEEE (2010)

\bibitem{colorization}
Zhang, R., Isola, P., Efros, A.A.: Colorful image colorization. In: European
  conference on computer vision (ECCV) (2016)

\bibitem{zhang2017split}
Zhang, R., Isola, P., Efros, A.A.: Split-brain autoencoders: Unsupervised
  learning by cross-channel prediction. In: Proceedings of the IEEE Conference
  on Computer Vision and Pattern Recognition (CVPR) (2017)

\end{thebibliography}

\clearpage
\thispagestyle{empty}
\begin{center}
\textbf{\Large Supplementary material}
\end{center}

\stepcounter{myequation}
\stepcounter{myfigure}
\stepcounter{mytable}
\stepcounter{mysection}
\makeatletter
\renewcommand{\theequation}{S\arabic{equation}}
\renewcommand{\thefigure}{S\arabic{figure}}
\renewcommand{\thetable}{S\arabic{table}}
\renewcommand{\thesection}{S\arabic{section}}

\section{Background}
\label{background}

We briefly discuss some of existing self-supervised learning approaches that have been used for the analysis in this paper. 

\textbf{RotNet:} Rotation prediction, proposed by Gidaris et al. \cite{rotnet}, has been one of the most successful pretext tasks for the learning of useful semantic representations.  In this approach, the network is trained to predict one of the $K$ rotations which was used for transforming the input image $x_i$. The authors found that $K=4$ with $\mathcal{T} = \{0^{\circ},90^{\circ},180^{\circ},270^{\circ}\}$ produced the best results. Every image $x_i$ is transformed using all four rotation transformations $x^{t_1}_i$,  $x^{t_2}_i$,  $x^{t_3}_i$ and  $x^{t_4}_i$, and the network is trained to predict $t_1$, $t_2$, $t_3$ and $t_4$, which are the rotation angles used for transforming $x_i$. The base encoder $f_\theta$ is trained by minimizing the following loss function $\mathcal{L}$:
\begin{equation}
\label{rotnet}
    \mathcal{L}_{RotNet} =  \frac{1}{B}\sum_{i=0}^{B-1}\frac{1}{K}\sum_{k=0}^{K-1} \ell_{CE}(M_\theta(x^{t_k}_i),t_k)
\end{equation}
Here, $M_\theta$ represents the network that takes as input rotated images $x^{t_k}_i$, and outputs the softmax predictions over the four possible rotation angles.

\textbf{SimCLR:} The work by Chen et al. \cite{simclr} presents a Simple Framework for Contrastive Learning of Visual Representations (SimCLR), which utilizes existing architectures such as ResNet \citep{resnet}, and avoids the need for specialized architectures and memory banks. SimCLR proposed the use of multiple data augmentations, and a learnable nonlinear transformation between representations and the contrastive loss to improve the effectiveness of contrastive learning. The authors find the following augmentations to be best suited for the contrastive learning task - random crop and resize, random color jitter and random Gaussian blur. These augmentations are applied serially to every image $x_i$ to generate two independent augmentations $x^{a_1}_i$ and $x^{a_2}_i$, which are considered as positives in the contrastive learning task. The $2(B-1)$ augmentations of all other images in a batch of size $B$ are considered as negatives. The network is trained by minimizing the normalized temperature-scaled cross entropy loss (NT-Xent) loss with temperature $T$ as shown in Eq.(\ref{simclr}). The cosine similarity between two vectors $a$ and $b$ is denoted as $\textrm{sim}(a,b)$. The overall network formed by the composition of the base encoder $f_\theta$ and the projection network $g_\theta$ is represented by $M_\theta$.
\begin{equation}
\label{simclr} \small{
\begin{aligned}
  \mathcal{L}_{SimCLR} = - \frac{1}{2B}\sum_{i=0}^{B-1} \sum_{m=1}^{2} \log \frac{ \exp(\textrm{sim}(M_\theta(x^{a_1}_i), M_\theta(x^{a_2}_i)) / T)  }{  \sum_{j=0}^{B-1} \sum_{l=1}^{2} \mathbbm{1}_{[j \neq i]} \exp(\textrm{sim}(M_\theta(x^{a_m}_i), M_\theta(x^{a_l}_j)) / T)}
  \end{aligned}}
\end{equation}

\textbf{BYOL:} While prior approaches relied on the use of negatives for training, Grill et al. \cite{byol} proposed Bootstrap Your Own Latent (BYOL), which could achieve state-of-the-art performance without the use of negatives. The two augmentations $x^{a_1}_i$ and $x^{a_2}_i$ are passed through two different networks - the base network $M_\theta$, and the derived network $M_\psi$ respectively. The weights of the base network are updated using back-propagation, while the weights of the derived network are obtained by computing a slow exponential moving average over the weights of the base network. The base network is trained such that the representation of $x^{a_1}_i$ at its output can be used to predict the representation of the $x^{a_2}_i$ at the output of the derived network, using a predictor network $P_\theta$. The symmetric loss that is used for training the base network is shown below:

\begin{equation}
\label{byol}
\begin{aligned}
    \mathcal{L}_{BYOL} = - \frac{1}{2B} \sum_{i=0}^{B-1} \textrm{sim}(P_\theta (M_\theta(x^{a_1}_i)), M_\psi(x^{a_2}_i)) + \textrm{sim}(P_\theta (M_\theta(x^{a_2}_i)), M_\psi(x^{a_1}_i))
\end{aligned}
\end{equation}

\section{Eliminating false positives in self-supervised learning}
\label{FP_supp}
As shown in Fig.\ref{fig:noise_learning}(b), two random augmentations of a given image may not always be similar to each other. The use of very small crops increases the likelihood of obtaining augmentations which may be unrelated to each other. This leads to false positives in instance-similarity based learning approaches. In Table-\ref{tab:FP_byol}, we use Grad-CAM \citep{gradcam} based saliency maps to select crops such that mean saliency score of the cropped image is greater than that of the full image. We describe this method in more detail below. 

\textbf{Mean-saliency based cropping:} We denote the saliency map of an image using $G(x)$, which is a probability map indicating the importance of each pixel in the image. In order to select rectangular crops having high saliency score, we first calculate the mean probability score $P(x)$ for an image $x$ of dimension $W\times H$ as follows: 

\begin{equation}
    P(x) = \dfrac{1}{W \cdot H} \sum_{i=0}^{W} \sum_{j=0}^{H} G_{i,j}(x)
\end{equation}

For selecting a rectangular crop from the image, we randomly sample the top left corner coordinates $(l,m)$, width $w$, and height $h$ from the valid range. These values can be used to obtain a rectangular crop $x^{a_1}$. We formulate the saliency score of the crop $x^{a_1}$ as follows:

\begin{equation}
    P(x^{a_1}) = \dfrac{1}{w \cdot h} \sum_{i=l}^{l+w} \sum_{j=m}^{m+h} G_{i,j}(x)
\end{equation}

The sampled crop is accepted only if $P(x^{a_1}) >  P(x)$. We repeatedly sample until a valid crop is found, and restrict to a maximum of 10 tries. If no valid crop is found, we use a random crop. We observe that 10 tries are sufficient to find valid crops in most cases and random cropping is used for very few images. 

\textbf{Computational Budget:} As shown in Table-\ref{tab:FP_byol}, with $50$ epochs of training, the accuracy on BYOL baseline is $63.64\%$, which increases to $66.72\%$ with the use of supervised saliency maps. However, this method assumes the availability of a network which is pre-trained on a relevant dataset, which may not always hold true. Hence, the computational budget for training this reference network needs to be considered too. We use fully supervised network trained for $90$ epochs as the reference model for generation of saliency maps. Therefore, the total budget for the BYOL baseline is $140$ epochs ($50+90$). As shown in Table-\ref{tab:imagenet2}, the accuracy obtained by training the BYOL baselines for $100$ epochs is $71.02\%$ which is $4.3\%$ higher than the model that is trained for $50$ epochs using saliency maps, with an effective training budget of $140$ epochs. This shows that while the use of saliency maps from a pre-trained network helps improve accuracy, it is not a practical option in cases where a model that is pre-trained on a related dataset is not available a priori.

\section{Details on Datasets}
\label{sec:datasets}
We present our analysis and results across the following datasets: CIFAR-$10$, CIFAR-$100$ \citep{krizhevsky2009learning} and ImageNet-100 \citep{cmc}, which is a $100$-class subset of ImageNet \citep{imagenet_cvpr09}. We do not present our main results on the full ImageNet dataset due to computational limitations. However, we show the scalability of our approach to ImageNet on a short training schedule of $30$-epochs. Details of these datasets are presented below: \\
\textbf{CIFAR-10:} CIFAR-$10$ \citep{krizhevsky2009learning} is a $10$ class dataset comprising of $50,000$ images in the training set and $10,000$ images in the test set. The dataset consists of RGB images of dimension $32\times 32$. The images in the train and test sets are equally distributed across all classes. \\
\textbf{CIFAR-100:} CIFAR-$100$ \citep{krizhevsky2009learning} dataset consists of $50,000$ images in the training set and $10,000$ images in the test set, equally distributed across $100$ classes. The dimensions and number of channels of images in CIFAR-$100$ is the same as CIFAR-$10$. \\
\textbf{ImageNet:} ImageNet \citep{imagenet_cvpr09} is a $1000$-class dataset consisting of around $1.2$ million images in the training set and $50,000$ images in the validation set. We consider the validation set as the test set, since the true test set is held private. The dataset consists of RGB images of dimension $224\times224$. \\
\textbf{ImageNet-100:} ImageNet-100 is a $100$-class subset of the ImageNet dataset. We consider the same $100$ class subset that was used by Tian et al. \cite{cmc}. 

\section{Details on Training hyperparameters}
\label{sec:hyp}
We consider the following baselines for our experiments: SimCLR \citep{simclr}, BYOL \citep{byol}, SimSiam \cite{simsiam} and SwAV \citep{swav}. Since these papers primarily demonstrate results on the ImageNet dataset, using larger architectures and longer training schedules, we perform extensive hyperparameter search to obtain strong results for the baselines on the datasets considered. We use the ResNet-18 \citep{resnet} architecture for all experiments other than the ImageNet-1k runs, where ResNet-50 was used. The dimension of features before the last fully-connected classification layer is $512$, which is smaller than that of ResNet-50, where the dimension is $2048$. We fix the batch size to be $512$ in all our experiments. We discuss details on hyperparameter tuning for obtaining strong baselines in Section-\ref{sec:tuning_baselines}, and describe the same for the proposed method in Section-\ref{sec:tuning_ours}. 

\subsection{Details on the Baseline Implementation}
\label{sec:tuning_baselines}
\vspace{0.2cm}

\textbf{SimCLR:} For the SimCLR \citep{simclr} baseline on CIFAR-10 and CIFAR-100, we perform a hyperparameter search for the learning rate, weight decay and the temperature used in the loss. We tune the learning rate in the range of 0.1 to 1 with a step size of 0.1, and the temperature in the range of 0.1 to 0.5 with a step size of 0.1. For weight decay we search over the range \{ $5 \times 10^{-4}$, $1 \times 10^{-4}$, $1 \times 10^{-5}$, $1 \times 10^{-6}$ \}. Finally, we use a learning rate of 0.5, weight decay of $1 \times 10^{-4}$ and a temperature of 0.2 for all our experiments. Following the official implementation \cite{simclr}, we use cosine learning rate schedule with a warm-up of $10$ epochs. For the projection head, we use a $2$ layer MLP with the hidden layer consisting of $512$ nodes. The output is a $128$-dimensional vector. We use batch normalization layers \citep{batchnorm} in the projection head. For ImageNet-100, we use the implementation and tuned hyperparameters from the repository \textit{solo-learn} \cite{turrisi2021sololearn}.

\textbf{BYOL:} For BYOL \citep{byol} baselines on CIFAR-10 and CIFAR-100, we perform a search for the learning rate and weight decay in the same manner as described in the paragraph above. Additionally we tune the momentum $\tau$ of the target network in BYOL \citep{byol} from the values \{ $0.8, 0.85, 0.9, 0.95, 0.99, 0.995, 0.999$ \}. Finally, we use a learning rate of 0.8 and weight decay of $1 \times 10^{-4}$ for CIFAR-10 and CIFAR-100. We tune the learning rate for ImageNet-100 in the range 0.4 to 0.7 with a step size of 0.1. We finally use a learning rate of 0.6 and a weight decay of $1 \times 10^{-4}$ for ImageNet-100. We use $\tau$ of 0.95, 0.85 and 0.95 for CIFAR-10, CIFAR-100 and ImageNet-100 respectively.

\textbf{SimSiam:} For the SimSiam \citep{simsiam} baselines, we use the implementation from the repository \cite{patrick-hua}, and perform a hyperparameter search for the learning rate, weight decay and the number of projection layers used in the loss. We tune the learning rate in the range of 0.03 to 0.1 with a step size of 0.01, and additionally try 0.2 as well. For weight decay we search over the range \{ $6 \times 10^{-4}$, $5 \times 10^{-4}$, $4 \times 10^{-4}$, $3 \times 10^{-4}$, $1 \times 10^{-4}$, $1 \times 10^{-5}$, $1 \times 10^{-6}$ \}. For the number of projection layers, we consider two values, 2 and 3. Finally, for CIFAR10, we use a learning rate of 0.07, weight decay of $4 \times 10^{-4}$ and number of projection layers as 2. For CIFAR100, we use a learning rate of 0.05, weight decay of $5 \times 10^{-4}$ and number of projection layers as 3. Following the official implementation \citep{simsiam}, we use the cosine learning rate schedule with a warm-up of $10$ epochs. For the projection head, the hidden layer is set to $2048$ nodes and output is a $2048$-dimensional vector. For the prediction head, the hidden layer has $512$ nodes and the output is again a $2048$-dimensional vector. We use batch normalization layers in the projection and prediction heads similar to the official implementation \citep{simsiam}.

\textbf{SwAV:} We use the code and hyperparameters from the official implementation \cite{swav}. For CIFAR-10, we search for the optimal number of prototypes over the values \{10, 30, 50, 70, 90, 100, 120, 150\}, $\epsilon$ over \{0.01, 0.03\} and queue over \{0, 38, 384\}. we finally set the number of prototypes to 100 without using a queue, and set $\epsilon$ to 0.03. Since CIFAR-10 images are small in size (32x32), we do not use the multi-crop strategy. We use the same settings for CIFAR-100 as well. For ImageNet-100, we scale the default number of prototypes from the official code \cite{swav} by a factor of 10 to 300, based on the scaling of number to classes from 1000 to 100. We use search for queue length in the range \{0, 384, 1920, 3840\} and set it to 384 finally. For the ImageNet-1k runs, we skip the use of multi-crop augmentations to speed up the training.

\subsection{Details on the Proposed Implementation}
\label{sec:tuning_ours}

We use the same hyperparameters as the respective baselines for the implementation of the proposed method, and additionally tune only the value of $\lambda$ (Eq.\ref{eq:ours}), which is the weighting factor used for the rotation loss. We use a 2 layer MLP for the rotation prediction task and use batch normalization for the hidden layer. For finding the best setting of $\lambda$, we tune for $1 / (2 \cdot \lambda)$ in the range 1 to 10 with step size of 1, and for $2 \cdot \lambda$ in the range 0 to 1 with a step size of 0.1.  In order to minimize computational overheads, we use the same value of $\lambda$ as ImageNet-100 on ImageNet-1k as well. 

For SimCLR, we use $2 \cdot \lambda$ as 1 for CIFAR-10 and CIFAR-100, and 0.1 for ImageNet-100. For BYOL, we use $1 / (2 \cdot \lambda$) as 5 for CIFAR-10 and CIFAR-100, and 6 for ImageNet-100. For SimSiam, we set the value of $2 \cdot \lambda$ to 0.1 for CIFAR-10 and 0.2 for CIFAR-100. For SwAV, we set the value of $2 \cdot \lambda$ to 0.5 for CIFAR-10 and CIFAR-100, and 0.1 for ImageNet-100 and ImageNet-1k.

\subsection{Training Details of Linear Evaluation}

The linear evaluation stage consists of training a linear classification layer on top of the frozen backbone network. We do not update the batch statistics in this stage. For linear evaluation on CIFAR-10 and CIFAR-100, we do not apply any spatial augmentations to the images during training. We use the SGD optimizer with momentum of 0.9. We train for 100 epochs with a batch size of 512. We use a learning rate of $1.0$ which is the best setting chosen from the range \{ 0.1, 0.5, 1.0, 1.5, 2.0 \}. The same settings are used for ImageNet-100 BYOL linear evaluation as well.

For SimSiam linear evaluation, we apply Random cropping and horizontal flipping. We use the SGD optimizer with momentum over 100 epochs using a batch size of 512, learning rate of 30.0 and momentum of 0.9, as recommended by the authors \cite{simsiam}. Cosine scheduler with decay is employed without any warmup for the training.

On ImageNet-100, we use the settings from the repository \textit{solo-learn} \cite{turrisi2021sololearn} for the linear evaluation of SimCLR \cite{simclr}. For linear evaluation of SwAV models on ImageNet-100 and ImageNet-1k, we use the settings from their official repository \cite{swav}, and use 30 epochs of training on ImageNet-1k.

We use the same hyperparameters for the linear evaluation of the proposed approach and the respective baselines.

\subsection{Training Details of Semi-supervised learning}

We follow the semi-supervised training settings from \citep{swav, pcl} for both $1\%$ and $10\%$ labels. Specifically, we train for 20 epochs with a batch size of 256. For the setting of $1\%$ labels, we use a learning rate of 0.02 for the backbone and 5.0 for the linear layer. For the setting of $10\%$ labels, we use a learning rate of 0.01 for the backbone and 0.2 for the linear layer. We decay the learning rates by a factor of 0.2 at epochs 12 and 16 in both the settings. We do not use weight decay during the training.

\section{Ablation Experiments}
\label{abaltion_supp}
In this section, we present additional experiments and results to highlight the significance of various aspects of the proposed method. 

\vspace{-0.5cm}
\begin{table}[h!]
\caption{\textbf{Rotation Angles:} Ablation experiments to show the impact of the rotation set ($\mathcal{T}$) used in the proposed approach. K-Nearest Neighbor (KNN) classification accuracy ($\%$) with K=200 and Linear evaluation accuracy ($\%$) on the CIFAR-$100$ dataset are reported for the baseline (BYOL \citep{byol}) and variations in the proposed approach (BYOL + rotation).}
% \vspace{-0.1cm}
\label{tab:rot_angle}
\centering
% \resizebox{1.0\linewidth}{!}{%
\begin{tabular}{lccc}
\toprule
Rotation Set ($\mathcal{T}$) & ~~$|\mathcal{T}|$~~ & ~~KNN~~ & ~~Linear~~\\
\midrule
$\phi$ (BYOL \citep{byol}) &0& 54.37 & 60.67 \\
$\{0^{\circ},180^{\circ}\}$ &2& 58.03 & 66.21 \\
$\{90^{\circ},270^{\circ}\}$ &2& 53.86 & 62.96 \\
$\{0^{\circ},90^{\circ}\}$ &2& 56.41 & 65.24 \\
$\{0^{\circ},270^{\circ}\}$ &2& 56.29 & 65.04 \\
$\{0^{\circ},90^{\circ},180^{\circ},270^{\circ}\}$ &4& \textbf{58.41} & 67.03 \\
$\{45^{\circ},135^{\circ},225^{\circ},315^{\circ}\}$ &4& 57.60 & 65.50 \\
$\{0^{\circ},45^{\circ}, ... ,270^{\circ}\,315^{\circ}\}$ &8& 57.54 & \textbf{67.25} \\
$\{0^{\circ},30^{\circ}, ... ,300^{\circ}\,330^{\circ}\}$~~~ &12& 55.43 & 63.61 \\

\bottomrule
\end{tabular}
% }
\vspace{-0.5cm}
\end{table}
\subsection{Impact of Variation in Rotation Angles}

In the proposed method, we transform every input image using a rotation transformation $t(.)$ which is randomly sampled from the set $\mathcal{T} = \{0^{\circ},90^{\circ},180^{\circ},270^{\circ}\}$. We present results by varying the number of rotation angles in the set $\mathcal{T}$ with BYOL \citep{byol} as the base approach in Table-\ref{tab:rot_angle}. While the use of $8$ rotation angles results in the best results, we use $4$ rotation angles (which results in marginally lower accuracy after linear evaluation) due to the simplicity of implementation, since rotation by multiples of $90^{\circ}$ does not require additional transformations such as cropping and resizing. The use of two rotation angles with $\mathcal{T} = \{0^{\circ},180^{\circ}\}$ leads to a drop of $0.82\%$ in linear evaluation accuracy when compared to the proposed method of using $4$ rotation angles. However, this setting is still $5.54\%$ better than the BYOL baseline. Therefore, the surprisingly simple task of predicting whether an image is in the correct orientation, or turned upside down is sufficient to boost the performance of the baseline method significantly. In the two-angle prediction task, excluding the $0^{\circ}$ rotation angle with $\mathcal{T} = \{90^{\circ},270^{\circ}\}$ leads to a significant drop of $3.25\%$ when compared to using $\mathcal{T} = \{0^{\circ},180^{\circ}\}$. We further note that using rotation transformations that are uniformly spaced ($\mathcal{T} = \{0^{\circ},180^{\circ}\}$) leads to better performance when compared to the use of $\mathcal{T} = \{0^{\circ},90^{\circ}\}$ or $\mathcal{T} = \{0^{\circ},270^{\circ}\}$. 

These experiments show that the level of difficulty of the auxiliary task plays a crucial role in the representations learned. The task should neither be too difficult ($12$ rotation angles), nor should it be too easy ($2$ rotation angles). Moreover, since the test images would have $0^\circ$ rotation angle, it helps to include this as one of the classes in $\mathcal{T}$. 

\vspace{-0.5cm}
\begin{table}[h!]
\caption{\textbf{Effect of number of layers shared with the Rotation Task:} Ablation experiments to show the impact of number of layers shared with the rotation task in the proposed approach. K-Nearest Neighbor (KNN) classification accuracy ($\%$) with K=200 and Linear evaluation accuracy ($\%$) on the CIFAR-$100$ dataset are reported for the baseline (BYOL \citep{byol}) and variations in the proposed approach (BYOL + rotation).}
\label{table:sharedlayers}
\centering
% \resizebox{1.0\linewidth}{!}{%
\begin{tabular}{lcc}
\toprule
\multicolumn{1}{c}{Layers shared with Rotation Task} & ~~~~KNN~~~~ & ~~~~Linear~~~~ \\
\midrule
None (BYOL \citep{byol} baseline) & 54.37 & 60.67 \\

First Convolutional layer ($f_\theta$) & 50.36 & 52.50 \\
~~~~+ Block - 1 ($f_\theta$) & 50.98 & 52.84 \\
~~~~+ Block - 2 ($f_\theta$) & 51.75 & 54.85 \\
~~~~+ Block - 3 ($f_\theta$) & 52.77 & 58.31 \\
~~~~+ Block - 4 ($f_\theta$) & 58.29 & 66.06 \\
~~~~+ Projection network ($g_\theta$) & \textbf{58.41} & \textbf{67.03} \\
\bottomrule
\end{tabular}
\vspace{-0.5cm}
% }
\end{table}

\subsection{Impact of Number of Shared Layers across Tasks}

In the proposed approach, we share the base encoder $f_\theta$ and the Projection network $g_\theta$ between the instance-similarity task and the rotation task. We perform experiments to study the impact of varying the number of shared layers between the two tasks. The results of these experiments on the CIFAR-$100$ dataset with BYOL \citep{byol} as the base method are presented in Table \ref{table:sharedlayers}. The ResNet-18 architecture consists of a convolutional layer followed by $4$ residual blocks. As an example, for the case where only Block-1 is shared between the two tasks, we replicate the remaining part of $f_\theta$ and $g_\theta$ separately for the the rotation task. Thus in this case, the rotation task only impacts Block-1 of the final base encoder $f_\theta$. As shown in Table \ref{table:sharedlayers}, increasing the number of shared blocks results in better performance. In fact, sharing only the first few layers leads to a degradation in performance when compared to the BYOL baseline. This indicates that the rotation task indeed helps improve the convergence of the overall network, and is not merely helping with learning better filters in the initial layers, as was the case in RotNet \citep{rotnet} training.

\vspace{-0.5cm}
\begin{table}[h!]
\caption{\textbf{Robustness to Image Augmentations:} Ablation experiments to show the impact of color jitter augmentation on the baseline (BYOL \citep{byol}) and proposed method (BYOL + Rotation). K-Nearest Neighbor (KNN) classification accuracy ($\%$) with K=200 and Linear evaluation accuracy ($\%$) on the CIFAR-$10$ dataset are reported. The proposed method is significantly more robust to the absence of color jitter augmentation.}
\centering
\vspace{0.1cm}
% \resizebox{1.0\linewidth}{!}{
\label{table:no_jitter}
\begin{tabular}{lll}
\toprule
 & \multicolumn{1}{l}{KNN} & \multicolumn{1}{l}{Linear} \\
\midrule
BYOL \citep{byol} & 86.56 & 89.30 \\
BYOL (without Color Jitter) & 82.21\Drop{4.35}  & 85.90 \Drop{3.40} \\
BYOL + Rotation & 89.80 & 91.89 \\
BYOL + Rotation (without Color Jitter)  & 88.52\drop{1.28} & 91.28 \drop{0.61} \\
\bottomrule
\end{tabular}
\vspace{-0.5cm}
% }
\end{table}

\subsection{Robustness to Image Augmentations}

BYOL \citep{byol} is known to be more robust to image augmentations when compared to contrastive learning methods such as SimCLR \citep{simclr}. The authors claim that although color histograms are sufficient for the instance-similarity task, BYOL is still able to learn additional semantic features for the image even without color jitter. We compare the impact of removing the color jitter augmentation on the baseline (BYOL) and the proposed approach (BYOL + Rotation) on CIFAR-$10$ dataset in Table-\ref{table:no_jitter}. We observe that addition of rotation task boosts the robustness to such augmentations even further. The absence of color jitter leads to a drop of $3.4\%$ in linear evaluation accuracy of BYOL, whereas the drop in accuracy for the proposed method without color jitter is only $0.61\%$, which is significantly lower. This makes the proposed method suitable for fine-grained image classification tasks as well, where the network needs to rely on color information for achieving good performance. 

\vspace{-0.5cm}
\begin{table*}[h!]
\caption{\textbf{Exploring Different Loss Formulations for the Rotation Task:} Ablation experiments to show the impact of different loss formulations on the rotation task. K-Nearest Neighbor (KNN) classification accuracy ($\%$) with K=200 and Linear evaluation accuracy ($\%$) on the CIFAR-$10$ dataset are reported. We additionally report the Rotation Task Accuracy ($\%$) obtained by freezing the base encoder $f_\theta$ and training a 2-layer MLP for the rotation classification task.}
\vspace{0.1cm}
\label{table:exploring_loss}
\centering
\resizebox{1.0\linewidth}{!}{%
\begin{tabular}{lccc}
\toprule
 & ~~KNN~~ & ~~Linear~~ & Rotation Acc ($f_\theta$) \\
\midrule
BYOL Baseline \citep{byol} & 86.56 & 89.30 & 73.40\\
\textbf{Ours} (Classification with CE Loss) & \textbf{89.80} & \textbf{91.89} &93.73\\
Classification with SupCon \citep{khosla2020supervised} Loss & 88.05 & 90.19& 81.86\\
Minimizing cosine similarity between Rotation Augmentations & 86.84 & 88.95&77.27 \\
BYOL + Rotation Augmentation & 74.32 & 79.70& 66.61 \\
Ours (BYOL + Rotation) + Rotation Augmentation  & 84.49 & 87.75& \textbf{94.24} \\
\bottomrule
\end{tabular}}
\vspace{-0.5cm}
\end{table*}
\subsection{Exploring Different Loss Formulations for the Rotation Task}

The proposed approach combines Cross-Entropy (CE) loss for the rotation task with various instance-similarity based tasks as shown in Eq.\ref{eq:ours}. We explore the use of different loss formulations for the rotation task with BYOL \citep{byol} as the base method on the CIFAR-$10$ dataset in Table-\ref{table:exploring_loss}. We first replace the CE loss for rotation with SupCon \citep{khosla2020supervised} loss, where all images with a similar rotation angle are treated as positives, while the remaining images in the batch are treated as negatives. This results in a significant drop of $1.7\%$ in the Linear evaluation accuracy. We observe a larger drop of $2.94\%$ when the CE loss is replaced with cosine similarity between two unique rotation augmentations sampled from the transformation set $\mathcal{T} = \{0^{\circ},90^{\circ},180^{\circ},270^{\circ}\}$. While the three approaches of minimizing CE loss, SupCon loss and cosine similarity between rotation augmentations seek to cluster similarly rotated images together and repel others, we find large differences in the representations learned. This shows that explicitly enforcing fixed categories in the auxiliary task helps in building a global semantic representation which is reinforced across training batches. This is exclusively achieved in the minimization of CE loss since it considers specific rotation based categories.  

We study the impact of adding rotations from the set  $\mathcal{T} = \{0^{\circ},90^{\circ},180^{\circ},270^{\circ}\}$ as augmentations in the BYOL training pipeline. Contrary to the proposed approach, this would encourage representations that are invariant to rotation. This leads to a large drop of $9.6\%$ when compared to the BYOL baseline. This is consistent with the observations by Chen et al. \cite{simclr} that rotation as an augmentation is not helpful in learning good representations. By including the rotation classification task in addition to this in the training objective, the accuracy improves by $8.05\%$, although it is still lower than the BYOL baseline due to the inclusion of rotation as augmentations which is contrasting to the rotation classification objective. 

We further compare the rotation sensitivity of representations at the output of the base encoder $f_\theta$. Similar to the experiments in Section-\ref{sec:rot_cov}, we freeze the network till the $f_\theta$ and train a rotation task classifier over this using a $2$-layer MLP head. We measure the rotation task accuracy, which serves as an indication to the rotation sensitivity of the base network. We observe that the trend in accuracy on the linear evaluation task is similar to the rotation task accuracy, indicating that rotation-covariant representations are better for downstream tasks. While the use of rotation augmentation along with rotation task prediction achieves a very high rotation accuracy, its performance on the contrastive task is only $64.34\%$, which is significantly lower than the baseline and the proposed methods (Ref. Table-\ref{tab:eval_no_noise}). Therefore, the accuracy on linear evaluation task is also lower than these methods.

\section{Reduction in Noise during training}
\label{noise_supp}
\begin{figure}
    \vspace{-0.5cm}
     \centering
    \includegraphics[width=0.5\linewidth]{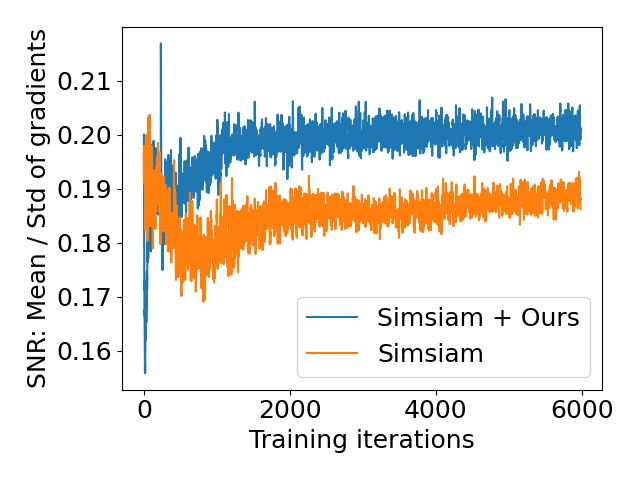}
    \caption{\small{Plot of SNR during SimSiam training on CIFAR-10 dataset. The use of rotation task (SimSiam+Ours) reduces the noise in the gradients, leading to faster convergence.}}
     \label{fig:snr}
    \vspace{-0.5cm}
 \end{figure}

To further demonstrate how the well-posedness of the rotation task reduces noise during the training, we plot the Signal-to-Noise ratio (SNR) of the gradients. For this, we follow Mitrovic et al. \cite{mitrovic2020icbinb} and at each iteration, we compute the ratio of the exponential moving average of mean and variance of the gradients, and average it across all the parameters to obtain the SNR. Fig-\ref{fig:snr} shows the progression of SNR during training on CIFAR-10 with SimSiam \cite{simsiam} as the base method. We find that our method indeed improves the SNR during training by providing a \textit{noise-free} supervisory signal and hence facilitates the learning of representations in an efficient and effective way.

\section{Transfer Learning}
\label{sec:transfer}

In this section, we report results using a ResNet-50 architecture with a 30-epoch training schedule. We perform the pretraining across 4 Nvidia Tesla V100 GPUs. We do not use multi-crop strategy in order to reduce the computational overheads. For all the ImageNet-1k runs, we do not perform additional hyperparameter tuning for the proposed approach, and use the same value of $\lambda$ that was best in the SwAV ImageNet-100 runs ($2\cdot \lambda = 0.1$). Using the linear evaluation training code and hyperparameters from the official SwAV repository for 30 epochs on the ImageNet-1k dataset, we achieve $54.9\%$ accuracy using the SwAV baseline, and $57.3\%$ accuracy using the proposed method, resulting in a gain of $2.4\%$ (Table-\ref{tab:transfer_aaai}). This shows that the proposed approach generalizes well to large-scale datasets and larger model capacities as well. 

\begin{table*}[h!]
\caption{\textbf{Transfer Learning (Classification):} Performance ($\%$) after linear evaluation on different datasets with a ResNet-50 backbone trained using SwAV \cite{swav} and the proposed approach.}
% \vspace{0.1cm}
\label{tab:transfer_aaai}
\resizebox{1.0\linewidth}{!}{%
\begin{tabular}{lccccccccccccc}
\toprule
            & ImageNet       & CIFAR-10       & CIFAR-100      & Flowers        & Caltech        & Aircraft       & DTD            & Cars           & Food           & Pets           & SUN            & VOC   & Avg         \\
            \midrule
SwAV \cite{swav}       & 54.90          & 86.22          & 64.18          & 83.53          & 80.91          & 38.78          & \textbf{69.79} & 31.65          & 59.41          & 70.73          & 52.48          & 76.33   & 64.08       \\
SwAV + Ours & \textbf{57.30} & \textbf{87.85} & \textbf{66.94} & \textbf{85.78} & \textbf{84.18} & \textbf{42.09} & 69.68          & \textbf{32.52} & \textbf{59.46} & \textbf{71.27} & \textbf{53.25} & \textbf{76.70} &\textbf{65.59} \\
\bottomrule
\end{tabular}}
\vspace{-0.5cm}
\end{table*}

\textbf{Classification:} We evaluate the generalization of the learned representations to other datasets by training a linear classifier on the pretrained backbone after freezing the weights of the backbone, as reported by Caron et al. \cite{swav}. We report transfer learning results on CIFAR-10 \cite{krizhevsky2009learning}, CIFAR-100 \cite{krizhevsky2009learning}, Oxford 102 Flowers \cite{nilsback2008automated}, Caltech-101 \cite{fei2004learning}, FGVC Aircraft \cite{maji2013fine}, DTD \cite{cimpoi2014describing}, Stanford Cars \cite{krause2013collecting}, Food-101 \cite{bossard2014food}, Oxford-IIIT Pets \cite{parkhi2012cats}, SUN397 \cite{xiao2010sun} and Pascal VOC2007 \cite{everingham2010pascal} datasets, as is common in literature \cite{kornblith2019better,simclr,ericsson2021well}. We use the code, hyperparameter tuning strategy and validation splits from the official repository of Ericsson et al. \cite{ericsson2021well} for obtaining results on the SwAV baseline. For the  evaluation of the proposed method, we use the best hyperparameters obtained for baselines, in order to highlight the gains obtained using the proposed approach more clearly. We achieve better performance across most of the datasets, and similar performance as the baseline on the DTD dataset \cite{cimpoi2014describing}. This is possibly because the DTD dataset is composed of textures only, and the images are rotation invariant. Therefore, learning representations that are covariant to rotation does not help in this case. Overall, we obtain an average improvement of $1.51\%$ across all datasets.

% \vspace{-0.5cm}
\begin{table}[h!]
\caption{\textbf{Transfer Learning (Object Detection):} Performance (AP, AP50 and AP75) on Pascal VOC \cite{everingham2010pascal} dataset for the task of Object Detection using Faster RCNN \cite{faster2015towards} FPN \cite{lin2017feature}  with a ResNet-50 backbone that is pretrained using SwAV \cite{swav} and the proposed approach. Pascal VOC07+12 trainval dataset is used for training and VOC07 test is used for evaluation. We consider two settings for evaluation: first with the ResNet-50 backbone being frozen, and second with the backbone being updated during training (Finetune).}
\vspace{0.1cm}
\centering
\label{tab:transfer_detection}
% \resizebox{1.0\linewidth}{!}{%
\begin{tabular}{lcccccc}
\toprule
\multirow{2}{*}{Method} & \multicolumn{3}{c}{VOC (Frozen)} & \multicolumn{3}{c}{VOC (Finetune)} \\

                        & AP        & AP50      & AP75     & AP         & AP50      & AP75      \\
                        \midrule
SwAV \cite{swav}                   & 44.10     & 74.54     & 45.00    & 43.80      & 74.46     & 45.07     \\
SwAV + Ours             & \textbf{45.12}     & \textbf{75.37}     &\textbf{46.67}    & \textbf{45.19}      & \textbf{75.17}     & \textbf{46.67} \\
\bottomrule
\end{tabular}
% }
% \vspace{-0.5cm}
\end{table}
\textbf{Object Detection: }
We evaluate the generalization of the learned representations to the task of Object Detection on the Pascal VOC dataset \cite{everingham2010pascal} using Faster RCNN \cite{faster2015towards} with Feature Pyramid Network \cite{lin2017feature} as the backbone. Pascal VOC07+12 trainval dataset is used for training and VOC07 test is used for evaluation. We consider two settings for evaluation: first with the ResNet-50 backbone being frozen, and second with the backbone being updated during training (Finetune). The training is done using the detectron2 framework \cite{wu2019detectron2} and their hyperparameters, as used by Ericsson et al. \cite{ericsson2021well}. As shown in Table-\ref{tab:transfer_detection}, we obtain consistent gains across the metrics AP, AP50 and AP75 in both evaluation settings.

\section{Additional Results on ImageNet}
\label{imagenet_supp}
In this section, we show additional results of the proposed method on ImageNet using the ResNet-50 architecture. Table-\ref{tab:longer_epochs} shows the model performance for longer training epochs, highlighting that the proposed approach can indeed scale to a longer training regime as well. We also achieve consistent gains of around $2.5\%$ over the SwAV baselines for 30 and 50 epoch runs with and without multicrop, and across different batch sizes as shown in Table-\ref{tab:wrt_swav}.
\vspace{-0.7cm}
\begin{table}[h]
\begin{minipage}{0.48\linewidth}
\caption{IN-1K: Improvements obtained on longer training epochs}
\vspace{0.1cm}
\setlength\tabcolsep{2pt}
\resizebox{1.0\linewidth}{!}{
\label{tab:longer_epochs}
\begin{tabular}{ccc}
\toprule
\textbf{\#Epochs} & \textbf{Method} & \textbf{Linear Acc (\%)} \\
\midrule
35                & SwAV + Ours & 59.5                \\ 
50                & SwAV + Ours & 61.9                \\ 
100                & SwAV + Ours& 64.3                \\
\bottomrule
\end{tabular}
}
 \vspace{-0.1cm}
\end{minipage}
\hfill
\begin{minipage}{0.48\linewidth}
\caption{IN-1K: Performance across different settings}
\vspace{0.1cm}
\setlength\tabcolsep{2pt}
\resizebox{1.0\linewidth}{!}{
\label{tab:wrt_swav}
\begin{tabular}{ccc}
\toprule
\textbf{Method}      & \textbf{30 ep, B256} & \textbf{50 ep, B1024} \\
\textbf{}      & \textbf{w/o Multicrop} & \textbf{with Multicrop} \\

\midrule
SwAV        & 54.9                & 65.8                   \\
SwAV + Ours & 57.3           & 68.3   \\
\bottomrule
\end{tabular}
}
\end{minipage}
\vspace{-0.7cm}
\end{table}

\end{document}